\documentclass[10pt,journal,compsoc]{IEEEtran}

\usepackage{amsmath,amssymb,amsfonts}

\usepackage{graphicx}
\usepackage{booktabs}
\usepackage{multirow}
\usepackage{array}

\usepackage{enumitem}
\usepackage{xspace}
\usepackage{microtype}

\usepackage{xcolor}
\usepackage{colortbl}

\usepackage{cite}
\usepackage{placeins}

\usepackage[hidelinks]{hyperref}

\providecommand{\method}{PhysioWave-T\xspace}

\newcolumntype{N}[1]{>{\centering\arraybackslash}p{#1}}
\newcolumntype{V}[1]{>{\raggedright\arraybackslash}p{#1}}

\newlength{\NSTMetricWidth}

\definecolor{NSTInk}{HTML}{171B22}
\definecolor{NSTGray}{HTML}{D8DEE8}
\definecolor{NSTBlue}{HTML}{C3D2E3}
\definecolor{NSTLavender}{HTML}{B9ACCF}
\definecolor{NSTRose}{HTML}{BE91B5}
\definecolor{NSTPeach}{HTML}{F1D5AE}
\definecolor{NSTIndigo}{HTML}{737BA3}
\definecolor{NSTBestInk}{HTML}{3C2853}
\definecolor{NSTMuted}{HTML}{353B48}
\definecolor{NSTDeltaInk}{HTML}{48233D}

\colorlet{NSTHead}{NSTBlue!65!white}
\colorlet{NSTSubhead}{NSTHead}
\colorlet{NSTGroup}{NSTGray!20!white}
\colorlet{NSTBaselineBlue}{NSTBlue!24!white}
\colorlet{NSTBaselineLilac}{NSTLavender!18!white}
\colorlet{NSTOurs}{NSTLavender!32!white}
\colorlet{NSTFocus}{NSTPeach!48!white}
\colorlet{NSTRule}{NSTIndigo!75!NSTInk}
\colorlet{NSTDropA}{NSTRose!16!white}
\colorlet{NSTDropB}{NSTRose!30!white}
\colorlet{NSTDropC}{NSTRose!46!white}
\colorlet{NSTDropD}{NSTRose!65!white}

\newcommand{\NSTBest}[1]{\textcolor{NSTBestInk}{\textbf{#1}}}
\newcommand{\NSTSecond}[1]{\textcolor{NSTInk}{\underline{#1}}}
\newcommand{\NSTVenue}[1]{\,{\fontsize{6.5}{7.5}\selectfont\textcolor{NSTMuted}{#1}}}
\newcommand{\NSTIndent}{\hspace{0.5em}}
\newcommand{\NSTGroupRow}[1]{%
  \addlinespace[2pt]%
  \rowcolor{NSTGroup}\multicolumn{6}{l}{%
  \rule{0pt}{2.2ex}\textcolor{NSTInk}{\textbf{#1}}}\\}
\newcommand{\NSTDropA}[1]{\cellcolor{NSTDropA}\textcolor{NSTDeltaInk}{\ensuremath{-#1}}}
\newcommand{\NSTDropB}[1]{\cellcolor{NSTDropB}\textcolor{NSTDeltaInk}{\ensuremath{-#1}}}
\newcommand{\NSTDropC}[1]{\cellcolor{NSTDropC}\textcolor{NSTDeltaInk}{\ensuremath{-#1}}}
\newcommand{\NSTDropD}[1]{\cellcolor{NSTDropD}\textcolor{NSTDeltaInk}{\ensuremath{-#1}}}
\newcommand{\NSTTableStyle}{%
  \color{NSTInk}%
  \arrayrulecolor{NSTRule}%
  \setlength{\heavyrulewidth}{0.65pt}%
  \setlength{\lightrulewidth}{0.3pt}%
  \setlength{\cmidrulewidth}{0.3pt}%
  \setlength{\aboverulesep}{0pt}%
  \setlength{\belowrulesep}{0pt}%
  \renewcommand{\arraystretch}{1.38}%
}


\begin{document}

\title{PhysioWave-T: Token-Efficient Wavelet Foundation Models for Biosignals}

\author{Yanlong~Chen,
        Simone~Benatti,
        Luca~Benini,~\IEEEmembership{Fellow,~IEEE,}
        Cuntai~Guan,~\IEEEmembership{Fellow,~IEEE,}
        and~Yawei~Li%
\IEEEcompsocitemizethanks{%
\IEEEcompsocthanksitem Y. Chen and Y. Li are with the School
of Electrical and Electronic Engineering, Nanyang Technological University,
Singapore 639798, and were previously with the Integrated Systems
Laboratory, ETH Z\"urich, 8092 Z\"urich, Switzerland. E-mail: \texttt{yanlong.chen@ntu.edu.sg}; \texttt{yawei.li@ntu.edu.sg}.
\IEEEcompsocthanksitem S. Benatti is with the Department of Engineering
``Enzo Ferrari'' (DIEF), University of Modena and Reggio Emilia,
41125 Modena, Italy. E-mail: \texttt{simone.benatti@unimore.it}.
\IEEEcompsocthanksitem L. Benini is with the Integrated Systems Laboratory,
ETH Z\"urich, 8092 Z\"urich, Switzerland, and also with the Department of
Electrical, Electronic and Information Engineering ``Guglielmo Marconi'' (DEI),
University of Bologna, 40136 Bologna, Italy. E-mail: \texttt{lbenini@iis.ee.ethz.ch}.
\IEEEcompsocthanksitem C. Guan is with the College of Computing and Data Science, Nanyang Technological University,
Singapore 639798. E-mail: \texttt{CTGuan@ntu.edu.sg}.

\IEEEcompsocthanksitem Corresponding author: Yawei Li.}%
}

\markboth{IEEE Transactions on Pattern Analysis and Machine Intelligence}%
{Chen \MakeLowercase{\textit{et al.}}: PhysioWave-T: Token-Efficient Wavelet Foundation Models for Biosignals}

\IEEEtitleabstractindextext{%
\begin{abstract}
We present \method, a token-efficient wavelet foundation-model family for
electroencephalography (EEG), electrocardiography (ECG), and surface
electromyography (sEMG). Dynamic scale folding recombines the $J{+}1$ wavelet
subbands before tokenization, compressing the sequence from $(J{+}1)CS$ to $CS$
tokens for $J$ decomposition levels, $C$ channels, and $S$ temporal patches.
Route-specific wavelet front ends and learned channel-name embeddings let one
Transformer per modality accommodate diverse channel counts and sampling rates,
while unseen configurations reuse the pre-trained backbone via a lightweight
front end. Together, these components scale PhysioWave to EEG: we pre-train a
30M-parameter EEG foundation model on over 15M windows spanning 19 to 128
channels, alongside ECG (30M) and sEMG (12M) encoders. Across four EEG routes,
folding cuts Transformer latency by 4.0--7.2$\times$ and peak memory by
6.8--12.7$\times$ over scale expansion without accuracy loss. The EEG encoder
leads eight of nine metrics against BENDR, BIOT, LaBraM, and EEGPT, and
surpasses LUNA-L on wearable seizure detection with fewer parameters.
Reliability-aware latent fusion (RALF) then integrates summary tokens from the
independently pre-trained encoders using per-sample reliability weights. On
SeizeIT2 at 11 false alarms/h, event sensitivity is 61.4\,\% for EEG alone,
74.4\,\% for RALF, and 78.9\,\% for weighted-logit fusion. RALF raises
window-level AUPRC from 0.321 (EEG alone) to 0.399. All six RALF variants
outperform weighted-logit and concatenation fusion on this metric. \method also
sets new state-of-the-art sEMG accuracies and the highest macro-F1 on PTB-XL
Super and Chapman--Shaoxing, establishing an efficient, extensible foundation
for biosignal representation learning. Code and models are available at
\url{https://github.com/ForeverBlue816/PhysioWave}.
\end{abstract}

}

\maketitle
\IEEEdisplaynontitleabstractindextext
\IEEEpeerreviewmaketitle

\section{Introduction}
\label{sec:introduction}
\IEEEPARstart{P}{hysiological} signals reflect the underlying dynamics of the
human body across diverse temporal scales, yet their acquisition hardware and
recording protocols are far from uniform. Across clinical and consumer settings,
EEG spans two-channel wearables to 128-channel scalp arrays; ECG ranges from
single-lead ambulatory patches to clinical twelve-lead systems; and sEMG is
captured by wristbands, forearm arrays, and high-density grids, each governed by
a distinct sampling rate and sensor geometry. A truly reusable representation
must capture multiscale physiological structure while flexibly accommodating this
hardware diversity, yet these two desiderata pull directly against each other.
Wavelet decompositions are well suited to non-stationary biosignals, but exposing
every scale to a Transformer incurs a steep quadratic cost: a $J$-level
decomposition of $C$ channels and $S$ patches yields $(J{+}1)CS$ tokens, causing
channel density, window duration, and decomposition depth to lengthen the token
sequence in tandem. Mapping every recording onto a single fixed input layout
sidesteps this variability, but intermediate heuristics such as channel
selection, padding, or resampling inevitably discard acquisition-specific
information along the way.

Our prior works, WaveFormer~\cite{chen2025waveformer} and
PhysioWave~\cite{chen2026physiowave}, learn multiscale representations of ECG
and sEMG through adaptive wavelet selection, soft gating, cross-scale feature
aggregation, and frequency-guided masked pre-training. Nevertheless, three
design choices in PhysioWave limit how far this paradigm scales. Its tokenizer
expanded every wavelet scale into a separate token row, and each encoder was
constrained to a single fixed input configuration. Both limitations are most
severe for EEG, whose clinical montages vary widely across datasets and whose
dense electrode arrays rapidly multiply the token count. PhysioWave's
multimodal experiments relied on an external pre-trained backbone and combined
modalities using fixed learned weights that could not adapt to the signal
quality of individual samples. These bottlenecks raise three central questions
that anchor this work: how to retain multiscale time-frequency information
without multiplying tokens, how to share a single architecture across
heterogeneous acquisition setups, and how to fuse modalities whose reliability
varies across recordings.

\method addresses the first two questions by decoupling acquisition-specific
processing from representation learning. Each acquisition route, defined by a
channel count and sampling rate, receives a dedicated wavelet front end and
sampling-rate-specific patch projection, while sharing a single Transformer
backbone and a name-based channel vocabulary across the modality. Within each
front end, dynamic scale folding recombines wavelet subbands prior to
tokenization using weights predicted from local subband statistics and refined
by learned synthesis filters. Consequently, the backbone processes $CS$ rather
than $(J{+}1)CS$ tokens regardless of decomposition depth: deeper analysis
enriches individual tokens instead of multiplying sequence length. Because
folding contains no channel-specific parameters, it transfers unchanged across
montages; an unseen downstream configuration requires only attaching a
lightweight front end to the frozen or fine-tuned backbone and appending any
new sensor names to the vocabulary.

These mechanisms allow PhysioWave to scale to EEG, yielding a unified EEG
foundation model as a central contribution of this work. Without coercing
recordings into a common montage, we pre-train a 30M-parameter encoder on over
15\,M four-second windows spanning clinical, research, brain--computer
interface, and high-density corpora (19--128 channels at 256 and 512\,Hz).
Across four EEG routes, scale folding reduces Transformer latency by
4.0--7.2$\times$ and peak memory by 6.8--12.7$\times$ over scale expansion,
cutting per-batch memory on 128-channel recordings from ${\sim}32$\,GB to
2.5\,GB. Against BENDR, BIOT, LaBraM, and EEGPT, our encoder leads on eight of
nine metrics across sleep staging, P300, and error-related negativity
detection---including on Sleep-EDFx, whose 100\,Hz derivations are unseen
during pre-training. On two-channel behind-the-ear seizure detection, it reaches
61.4\,\% sensitivity at ${\leq}11$ false alarms per hour, outperforming LUNA-L
with 25.4\,M versus 40.4\,M parameters. Pre-training also boosts label
efficiency: using just 10\,\% of KaggleERN labels, the encoder surpasses a
fully supervised scratch model. Finally, the learned representations capture
intrinsic physiology: channel embeddings recover scalp topography without
coordinate supervision, and the slowest wavelet scale's weight progressively
increases from wakefulness to deep sleep.

For the third question, reliability-aware latent fusion (RALF) combines this EEG
encoder with independently pre-trained ECG and sEMG encoders, without joint
multimodal pre-training. Each encoder contributes four summary tokens. A
per-sample reliability score gates that modality's contribution to attention
and the residual unimodal prediction, and missing modalities are masked
explicitly. On SeizeIT2, adding ECG and EMG to wearable EEG raises event
sensitivity at 11 false alarms/h from 61.4\,\% to 74.4\,\% with RALF and
78.9\,\% with weighted-logit fusion. RALF improves window-level AUPRC from
0.321 (EEG alone) to 0.399. All six variants outperform weighted-logit and
concatenation fusion in AUPRC, most clearly when ECG is included. RALF also
retains an AUPRC of at least 0.33 under every tested single-modality
corruption. Relative to PhysioWave, the contributions are:
\begin{itemize}[leftmargin=1.4em]
  \item \textbf{Token-efficient multiscale modeling.} Dynamic scale folding
  shortens the sequence by a factor of $J{+}1$ while retaining adaptive wavelet
  analysis. We quantify its effects on accuracy, latency, memory, and
  computation.
  \item \textbf{An EEG foundation model across acquisition configurations.}
  Route-specific front ends and extensible channel vocabularies let one shared
  30M-parameter backbone learn from heterogeneous montages and transfer to
  unseen ones, including two-channel wearable EEG.
  \item \textbf{Reliability-aware multimodal prediction.} RALF fuses compact
  modality summaries using sample-specific reliability estimates. We evaluate
  it on seizure detection under complete, missing, and corrupted inputs.
\end{itemize}

\begin{figure*}[htbp]
  \centering
  \includegraphics[width=\textwidth]{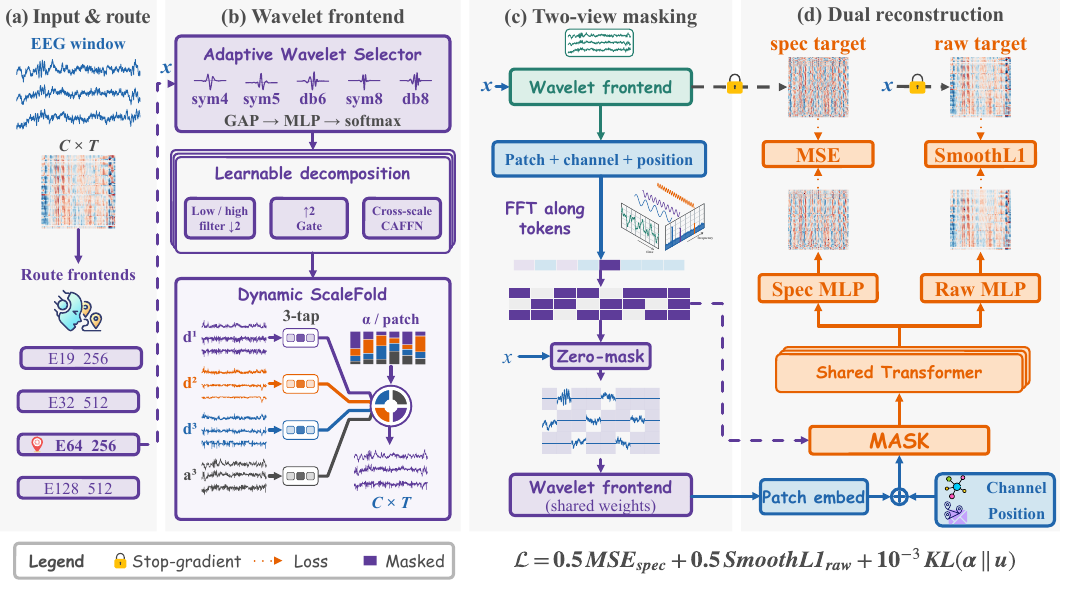}
  \caption{Overview of \method, shown for EEG. (a)~Channel count and sampling
  rate select a route-specific front end. (b)~Adaptive wavelet decomposition and
  dynamic scale folding produce one $C\times T$ signal. (c)~Spectral importance
  of clean-view tokens selects masked patches, zeroed before the online front
  end. (d)~A shared Transformer and two MLP heads reconstruct raw and
  folded-wavelet patches against stop-gradient targets.}
  \label{fig:overview}
\end{figure*}

\section{Related Work}
\label{sec:related}

\subsection{Physiological Foundation Models}
Following BERT~\cite{devlin2019bert} and MAE~\cite{he2022masked}, masked
reconstruction and predictive learning have produced transferable EEG
representations. BENDR~\cite{kostas2021bendr} and Brant~\cite{zhang2023brant}
established contrastive and predictive pre-training. BIOT tokenizes
channel-wise spectral patches~\cite{yang2023biot}, LaBraM predicts discrete
neural codes~\cite{jiang2024large}, EEGPT combines masked reconstruction with
spatiotemporal alignment~\cite{wang2024eegpt}, and CBraMod separates spatial and
temporal interactions through criss-cross attention~\cite{wang2025cbramod}.
Beyond EEG, ST-MEM~\cite{na2024guiding}, HuBERT-ECG~\cite{coppola2024hubert},
and ECG-FM~\cite{mckeen2025ecg} apply masked modeling or self-distillation to
ECG, and large wristband datasets support sEMG models for movement and
interaction~\cite{salter2024emg2pose,kaifosh2025generic}. Generic time-series
models such as MOMENT~\cite{goswami2024moment} and UniTS~\cite{gao2024units}
pursue broader transfer. These models differ mainly in their tokenization and
pre-training targets. \method instead addresses the interface between
acquisition and representation: configuration-specific processing precedes a
shared token space, and multiscale analysis does not lengthen the backbone
sequence.

\subsection{Multiscale Representations and Heterogeneous Inputs}
Wavelets localize signal structure in time and
frequency~\cite{mallat1989theory,daubechies1992ten}. Best-basis
selection~\cite{coifman1992entropy} and coefficient
shrinkage~\cite{donoho1995noising} adapt this representation to signal
structure or noise, and neural wavelet operators make its filters
trainable~\cite{li2020wavelet}. WaveFormer~\cite{chen2025waveformer} introduced
soft wavelet selection for sEMG, and PhysioWave~\cite{chen2026physiowave} paired
learnable wavelet analysis with frequency-guided pre-training for ECG and sEMG.
However, PhysioWave allocated a separate token row to every band. \method
retains these analysis components but folds the subbands before tokenization,
so decomposition depth no longer lengthens the sequence. Generalizing classical best-basis selection, the fold replaces discrete wavelet-packet tree search with a learned, continuous scale mixture.

Heterogeneous acquisition poses a separate challenge. Perceiver
architectures compress variable inputs into latent
arrays~\cite{jaegle2021perceiver}, mixture-of-experts models route tokens
among specialized subnetworks~\cite{shazeer2017outrageously,fedus2022switch},
and channel-independent designs process each channel
separately~\cite{nie2022time}. In EEG, name-based embeddings identify
electrodes~\cite{jiang2024large,wang2024eegpt}, graph models encode spatial
adjacency~\cite{demir2021eeg}, and LUNA~\cite{doner2026luna} maps variable
electrode sets to a fixed set of latent queries. \method selects a front end
deterministically from acquisition metadata rather than learning token-level
routing. It also extends name-based identity to ECG leads and sEMG electrode
sites, adding vocabulary entries during downstream adaptation.

\subsection{Multimodal Physiological Fusion}
Combining biosignals improves prediction when modalities carry complementary
information, yet acquisition failures rarely affect all sensors equally.
Even available modalities may differ substantially in their signal quality.
Latent and bottleneck fusion restrict cross-modal interaction to compact
summaries~\cite{jaegle2021perceiver,nagrani2021attention}, avoiding attention
over every unimodal token. RALF follows this design and adds a sample-specific
reliability estimate for each modality, supervised by corruption severity,
while explicit masks handle absent inputs.

\section{PhysioWave-T Architecture}
\label{sec:method}
\method is built on a simple principle: acquisition-specific processing
happens before tokenization, and representation learning happens after it.
Route-specific front ends absorb differences in channel count and sampling
rate, mapping every recording into a common token space in which one
Transformer per modality learns shared physiological structure. These front
ends retain PhysioWave's learnable wavelet analysis~\cite{chen2026physiowave}
but differ in two respects: they fold the wavelet scales into one signal before
tokenization, and they attach an explicit identity to every channel. The first
change keeps the sequence short, and the second lets the same backbone
recognize sensors across configurations.

Concretely, Fig.~\ref{fig:overview} traces a window
$X\in\mathbb{R}^{C\times T}$, with $C$ channels and $T$ samples, through the
encoder. The front end first decomposes each channel into $J$ detail subbands
and one approximation, then folds these $J{+}1$ bands back into a single signal
$\tilde{X}\in\mathbb{R}^{C\times T}$ with the shape of the input.
Fixed-duration patches of $\tilde{X}$ are projected into tokens, to which
learned channel-name embeddings and sinusoidal position embeddings are added.
With $S$ patches per channel, the Transformer receives $C\,S$ tokens, whereas
PhysioWave's scale expansion produces $(J{+}1)\,C\,S$. Decomposition depth
therefore enriches the content of each token without lengthening the sequence,
an advantage that grows with the number of channels and matters most for dense
EEG arrays. The following subsections describe each stage in turn.

\subsection{Routes and Modular Front Ends}
\label{sec:method:routes}
Physiological corpora rarely share a single input format, and forcing them into
one would require dropping channels, padding, or resampling. \method instead
groups recordings into \emph{routes}. A route $r=(C_r,f_r)$ is defined by a
channel count and a sampling rate, both read from recording metadata rather
than predicted. Components tied to these quantities remain
specific: each route has a wavelet front end $\mathcal{F}_r$ whose per-channel
analysis filters and gates can specialize to the sensors of that configuration,
and each sampling rate has a patch embedding $\mathcal{P}_f$. Components that
should generalize across configurations, namely the channel vocabulary, the
position code, and the Transformer, are shared by all routes of a modality:
\begin{equation}
  Z = \mathrm{Transformer}\big(\mathcal{P}_{f_r}(\mathcal{F}_r(X)) + E_{\mathrm{ch}} +
  E_{\mathrm{pos}}\big).
  \label{eq:route}
\end{equation}
Defining patches by physical duration $\tau$ ensures that each token spans the same temporal window across all sampling rates, aligning tokens from heterogeneous routes onto a common temporal footing. During
pre-training, the routes cover 19--128 EEG channels at 256 or 512\,Hz, 1--12
ECG leads at 250 or 500\,Hz, and 16--64 sEMG channels at 2\,kHz; the pretrained corpus is listed in Table~\ref{tab:pretrain_sets}.

This architectural separation also governs how the encoder adapts after
pre-training.
Because the front end is fully convolutional, it accepts any window whose duration is
an integer multiple of the patch duration. A downstream task sampled at a
pre-training rate reuses the corresponding patch embedding, and if it also
matches a route's channel count and channel-slot convention, it reuses that
route's front end as well. Any other configuration receives a newly
instantiated front end and patch embedding for its $(C,f)$, attached to the
transferred backbone. In both cases, channel names supply the sensor identities
that connect the new input to what the backbone has already learned
(Section~\ref{sec:method:tokens}).

\begin{figure}[tbp]
  \centering
  \includegraphics[width=\columnwidth]{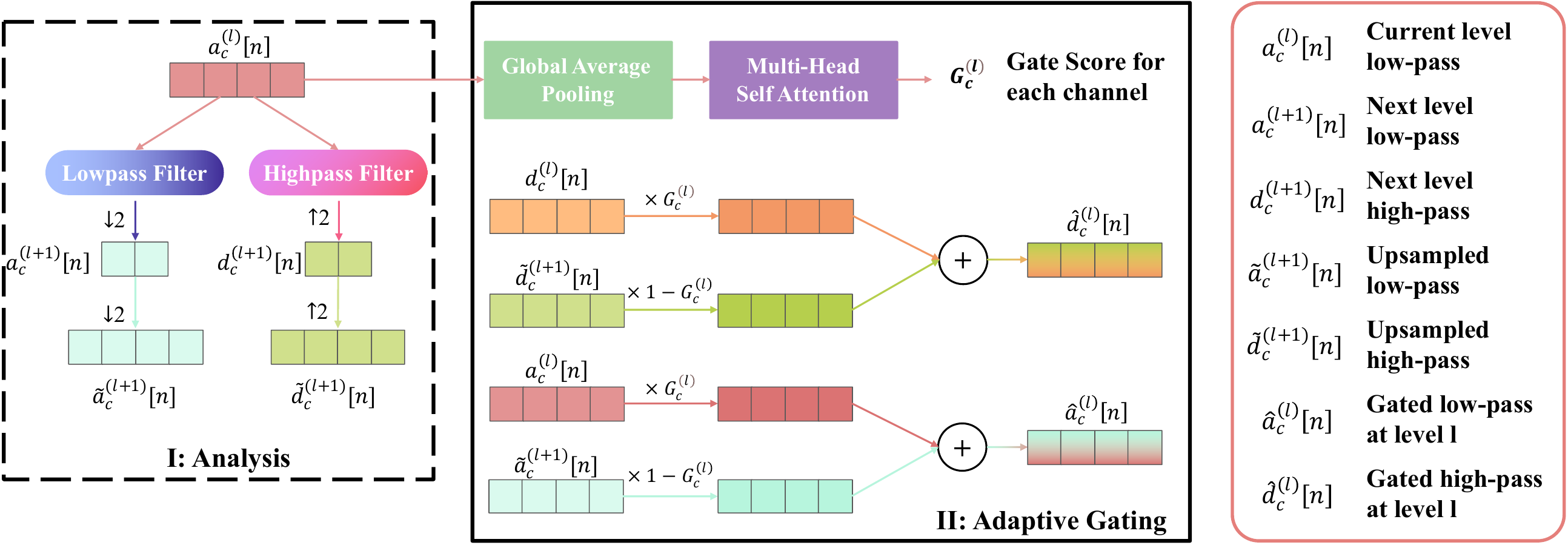}
  \caption{Soft-gated wavelet analysis. Channel-specific gates blend the upsampled subbands with the running approximation and detail states.}
  \label{fig:softgate}
\end{figure}

\subsection{Learnable Wavelet Decomposition}
\label{sec:method:wavelet}
The analysis stage follows PhysioWave and is summarized here to establish the notation used for folding. Each decomposition level combines adaptive basis
selection, analysis with soft gating, and cross-scale feature fusion, whose outputs form the subbands later recombined by scale folding.

\subsubsection{Adaptive Wavelet Selector}
The front end maintains $M$ learnable analysis-filter pairs
$\{(k_m^{\mathrm{lo}},k_m^{\mathrm{hi}})\}_{m=1}^{M}$, each with $K$ taps per
channel, so that no single wavelet family must fit every signal. The pairs are
initialized from named discrete wavelets, such as \texttt{db6} and
\texttt{sym4}, resampled to length $K$ and $\ell_1$-normalized. At level
$\ell$, a shared selector pools the current approximation over time, and a
compact MLP followed by a softmax predicts input-dependent mixture weights:
\begin{align}
  \pi^{(\ell)} &= \mathrm{Softmax}\big(\mathrm{MLP}(\mathrm{AvgPool}(a^{(\ell)}))\big)
  \in\mathbb{R}^{M},
  \label{eq:selector}\\
  k^{p,(\ell)} &= \sum_{m=1}^{M}\pi_m^{(\ell)}\,k_m^{p},
  \qquad p\in\{\mathrm{lo},\mathrm{hi}\}.
  \label{eq:mix}
\end{align}
The mixed filters serve as the low- and high-pass filters for analysis at that
level. The selector and filter bank share parameters across levels, while the
mixture weights adapt to the current approximation, allowing level-specific
basis choices. The taps are optimized jointly with the rest of the encoder
during pre-training. We reserve $\alpha$ for the distinct scale-folding weights
introduced in Section~\ref{sec:method:fold}.

\subsubsection{Analysis and Soft Gating}
Let $\downarrow_2$ and $\uparrow_2$ denote decimation and nearest-neighbor
upsampling by two. Each stage halves the temporal resolution while retaining
low- and high-frequency content. Starting from $a^{(0)}=X$ and $d^{(0)}=0$,
each channel is filtered and decimated into the next level:
\begin{equation}
\begin{aligned}
  a_c^{(\ell+1)} &= \downarrow_2\!\big(a_c^{(\ell)} * k^{\mathrm{lo},(\ell)}\big),\\
  d_c^{(\ell+1)} &= \downarrow_2\!\big(a_c^{(\ell)} * k^{\mathrm{hi},(\ell)}\big).
\end{aligned}
  \label{eq:analysis}
\end{equation}
The subbands are upsampled to the current resolution and blended through a
per-channel gate $G_c^{(\ell)}\in[0,1]$ (Fig.~\ref{fig:softgate}). Multi-head
attention over pooled channel statistics, followed by a sigmoid, predicts this
gate:
\begin{align}
  \hat{a}_c^{(\ell)} &= G_c^{(\ell)}\,a_c^{(\ell)} +
  \big(1-G_c^{(\ell)}\big)\uparrow_2\!a_c^{(\ell+1)},\\
  \hat{d}_c^{(\ell)} &= G_c^{(\ell)}\,d_c^{(\ell)} +
  \big(1-G_c^{(\ell)}\big)\uparrow_2\!d_c^{(\ell+1)}.
  \label{eq:gate}
\end{align}
Unlike a fixed skip connection, this blending learns the trade-off between
current-level context and decomposed features, mitigating artifacts from naive
upsampling. Through channel-specific weights, the front end modulates this contribution per channel and input on the fly.

\begin{figure}[tbp]
  \centering
  \includegraphics[width=0.8\columnwidth]{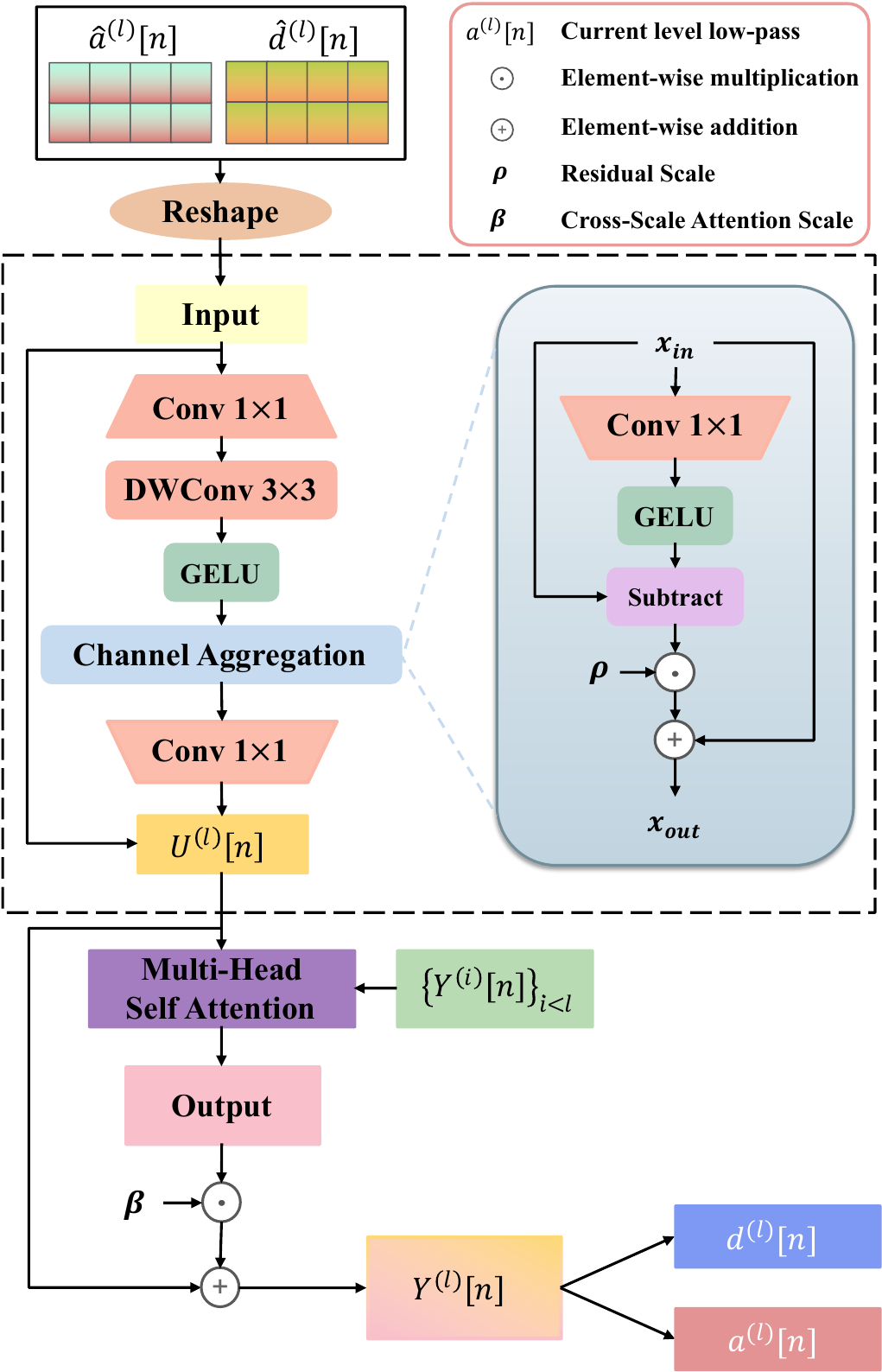}
  \caption{Cross-scale channel-aggregation feed-forward network (CAFFN). The current approximation--detail pair is refined through convolution, channel aggregation, and attention to shallower-level features.}
  \label{fig:caffn}
\end{figure}

\subsubsection{Cross-Scale Feature Fusion}
At each level $\ell$, the gated approximation and detail features are
concatenated along the channel axis into a $2C\times T$ map and processed by a
cross-scale channel-aggregation feed-forward network (CAFFN;
Fig.~\ref{fig:caffn}). CAFFN first refines this map locally: pointwise and
depthwise convolutions mix information across the approximation--detail pair
and along time, and channel aggregation with a learnable residual scale $\rho$
produces $U^{(\ell)}$. The current level then attends to the refined maps of
all shallower levels, with $U^{(\ell)}$ providing the queries and
$\{Y^{(i)}\}_{i<\ell}$ the keys and values:
\begin{equation}
  Y^{(\ell)} = U^{(\ell)} + \beta\,\mathrm{Attention}\big(U^{(\ell)},
  \{Y^{(i)}\}_{i<\ell}\big),
  \label{eq:caffn}
\end{equation}
where the learnable scalar $\beta$ balances current-level information against
finer-scale context. This fusion lets each subband draw on patterns already captured at finer resolutions, transforming otherwise isolated bands into cohesive, scale-aware features.

The refined map $Y^{(\ell)}$ is then split back into its approximation and
detail halves, $a^{(\ell)},d^{(\ell)}\in\mathbb{R}^{C\times T}$, and
decomposition continues from the approximation. After $J$ levels, the detail
subbands of every level and the final approximation are collected on a common
temporal grid:
\begin{equation}
  \mathrm{Spec}(X) = \big[d^{(1)}, d^{(2)}, \ldots, d^{(J)}, a^{(J)}\big]
  \in \mathbb{R}^{(J+1)C\times T}.
  \label{eq:spec}
\end{equation}
Each block of $C$ rows in $\mathrm{Spec}(X)$ thus carries one frequency band
across all channels. PhysioWave tokenized this $(J{+}1)C\times T$ map directly;
\method instead recombines its bands before tokenization, as described next.

\subsection{Dynamic Scale Folding}
\label{sec:method:fold}
Directly tokenizing $\mathrm{Spec}(X)$ creates a separate sequence for every
subband of every channel, multiplying the token count by $J{+}1$ and making
dense EEG arrays expensive. Dynamic scale folding instead recombines these
subbands before tokenization. For $b_s\in\mathbb{R}^{C\times T}$, the $s$-th
subband, $s=1,\ldots,J{+}1$, the folded representation is
\begin{align}
  \bar{b} &= \tfrac{1}{J+1}\sum_{s=1}^{J+1} b_s,
  \label{eq:mean}\\
  \tilde{X} &= \bar{b} + \gamma\Big(\sum_{s=1}^{J+1}\alpha_s\odot\big(g_s * b_s\big)
  - \bar{b}\Big).
  \label{eq:fold}
\end{align}
Here, $g_s$ is a short depthwise synthesis filter initialized to a delta, and
$\gamma$ is a learnable scalar initialized at $0.1$. The weights satisfy
$\alpha_s\in[0,1]$ and $\sum_s\alpha_s=1$. The mean $\bar{b}$ provides a reference
representation, with the residual branch learning scale-dependent corrections.
Delta-initialized filters and uniform weights make the initial fold an exact
average of the subbands, a neutral starting point.

The weights are predicted for each channel and each time block of one patch
duration. Each subband is summarized by log root-mean-square amplitude, the
entropy of its normalized squared amplitudes, and the logarithm of one plus its
kurtosis. These statistics are normalized across scales and passed through a
shared two-layer perceptron with 16 hidden units and a GELU activation. Its logits
are combined with a learnable scale prior:
\begin{equation}
  \alpha = \mathrm{Softmax}\big(\theta + \mathrm{MLP}(\mathrm{stats}(b_{1:J+1}))\big).
  \label{eq:alpha}
\end{equation}
The output layer and scale-prior logits start at zero, yielding uniform initial
weights. This soft selection is related to best-basis selection~\cite{coifman1992entropy},
whose entropy-based criterion is fixed; here, the criterion is learned.
A $\mathrm{KL}(\alpha\,\|\,u)$ penalty toward the uniform distribution, weighted
by $10^{-3}$, discourages concentration on a single band. The predictor is shared
across channels, and $g_s$, $\theta$, and $\gamma$ have no channel index. Folding
therefore introduces no channel-shaped parameters and can be transferred across
montages.

Folding reduces the token count from $(J{+}1)\,C\,S$ to $C\,S$, a fourfold
reduction for the three-level EEG decomposition. At fixed model width, the
quadratic attention term is consequently reduced by $(J{+}1)^2$; end-to-end
savings also depend on linear projections and front-end computation. Because
$\tilde{X}$ retains the shape of $X$, the front end remains a replaceable
interface to the shared backbone.

\subsection{Tokenization and Sequence Modeling}
\label{sec:method:tokens}
\label{sec:method:backbone}
The folded signal is divided into $S=T/(\tau f_r)$ patches per channel, and the
sampling-rate-specific embedding $\mathcal{P}_{f_r}$ projects each patch to
width $D$. The result is a channel-major sequence of $C\,S$ tokens, independent
of decomposition depth, in which token $c\,S+p$ corresponds to channel $c$ and
temporal patch $p$.

\textbf{Channel identity.} Because routes differ in which sensors occupy which
input slots, each token must also carry the identity of its sensor. Each
modality therefore maintains a vocabulary of sensor names, covering EEG
electrodes, ECG leads, and sEMG sites on wristbands, forearm arrays, and grids.
A learned table $E\in\mathbb{R}^{|V|\times D_c}$ assigns each name an
embedding, whose normalized projection is added to every token of that channel:
\begin{equation}
  E_{\mathrm{ch}}[c] = \tanh(g)\,W_{\mathrm{ch}}\,\mathrm{Norm}\big(E[\mathrm{name}(c)]\big).
  \label{eq:channel}
\end{equation}
The scalar gate $g$ starts at zero, so channel identity enters gradually during
training. A name unseen during pre-training receives a new row, initialized from
the unknown-name embedding and learned during downstream adaptation, while known
names keep their pre-trained embeddings. Input slots follow the route
convention, ensuring that per-channel filters, name embeddings, and positional
codes refer to the same sensor.

\textbf{Position and backbone.} A two-dimensional sinusoidal code over channel
and patch indices is added to the tokens, and rotary position
embeddings encode relative offsets inside each attention
layer. The full $C\,S$ sequence is processed by a stack of pre-normalized
Transformer blocks with RMS
normalization~\cite{zhang2019root}, query--key normalization, and gated
feed-forward units. The resulting representations $Z$ serve pre-training,
downstream prediction, and multimodal fusion alike, and the backbone and
channel vocabulary are transferred across all tasks of a modality.

\section{Pre-training, Adaptation, and Multimodal Fusion}
\label{sec:training}
The EEG, ECG, and sEMG encoders share one pre-training objective but are
trained independently, each on the corpora of its own modality. A pre-trained
encoder can then be adapted to a downstream configuration or combined with the
other two through reliability-aware latent fusion.

\subsection{Self-Supervised Pre-training}
\label{sec:training:pretrain}

Pre-training extends PhysioWave's masked reconstruction in two ways: masking
acts on the raw input, and the encoder reconstructs two complementary targets
(Fig.~\ref{fig:overview}(c)--(d)). Masking tokens alone would not suffice for
this front end, because convolution and cross-scale fusion mix neighboring
samples before tokenization, leaving hidden content accessible through adjacent
features. Each window is therefore processed in two views. The clean view
supplies reconstruction targets and mask-selection scores, while the online
view zeros the selected raw-signal patches before the front end, so their
samples never enter the online features. After tokenization, the masked
positions are replaced by a learned mask token. Patch selection follows
PhysioWave's frequency-guided strategy: spectral importance $e$ is computed from
the summed magnitude spectrum along the clean-view token sequence and combined
with uniform noise $z$ as $s=0.6\,e+0.4\,z$. The highest-scoring 70\,\% of
valid tokens are masked, so spectrally dominant patches are hidden most often
and must be inferred from context. For ECG, the six limb leads within a
temporal patch are masked together, since their linear dependencies would
otherwise leak the hidden content, and precordial leads are masked
individually. Unavailable channel slots are excluded from both mask selection
and reconstruction scoring.

\begin{figure}[htbp]
  \centering
  \includegraphics[width=\columnwidth]{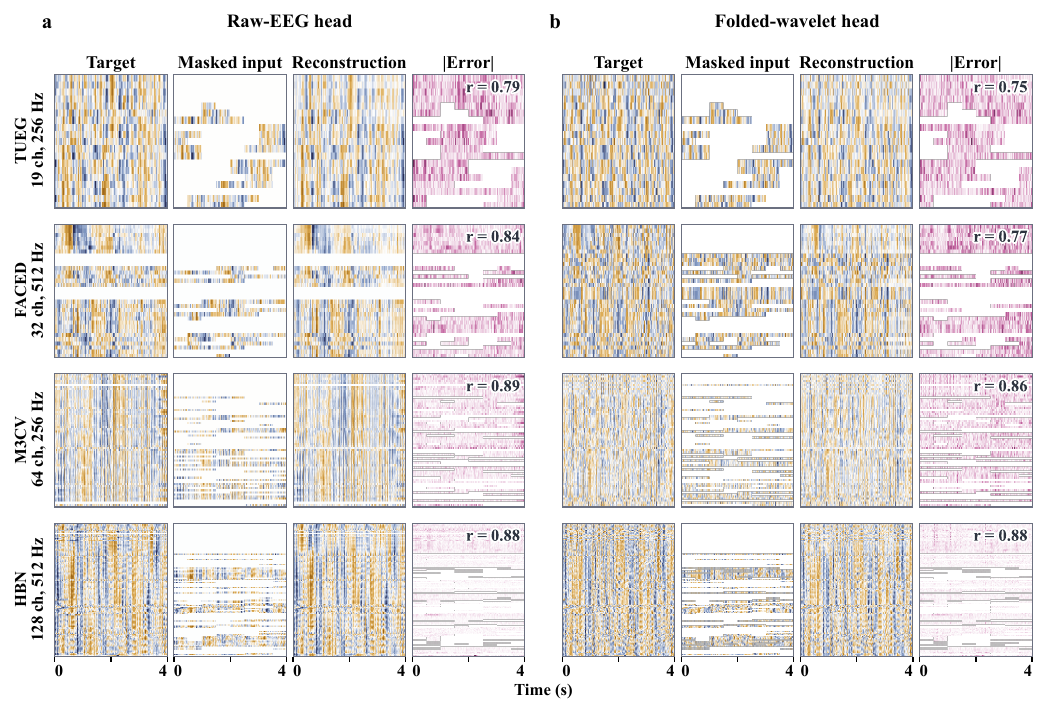}
  \caption{Masked reconstruction on four EEG routes spanning 19--128 channels,
  with 70\,\% of patches masked. (a)~Raw-EEG head. (b)~Folded-wavelet head.
  Columns show the target, masked input, reconstruction, and absolute error on
  masked patches; $r$ is the masked-patch correlation. Each row shows the
  median-NMSE validation window of its route, with target-derived color limits.}
  \label{fig:mask_reconstruction}
\end{figure}

\begin{figure}[htbp]
  \centering
  \includegraphics[width=\columnwidth]{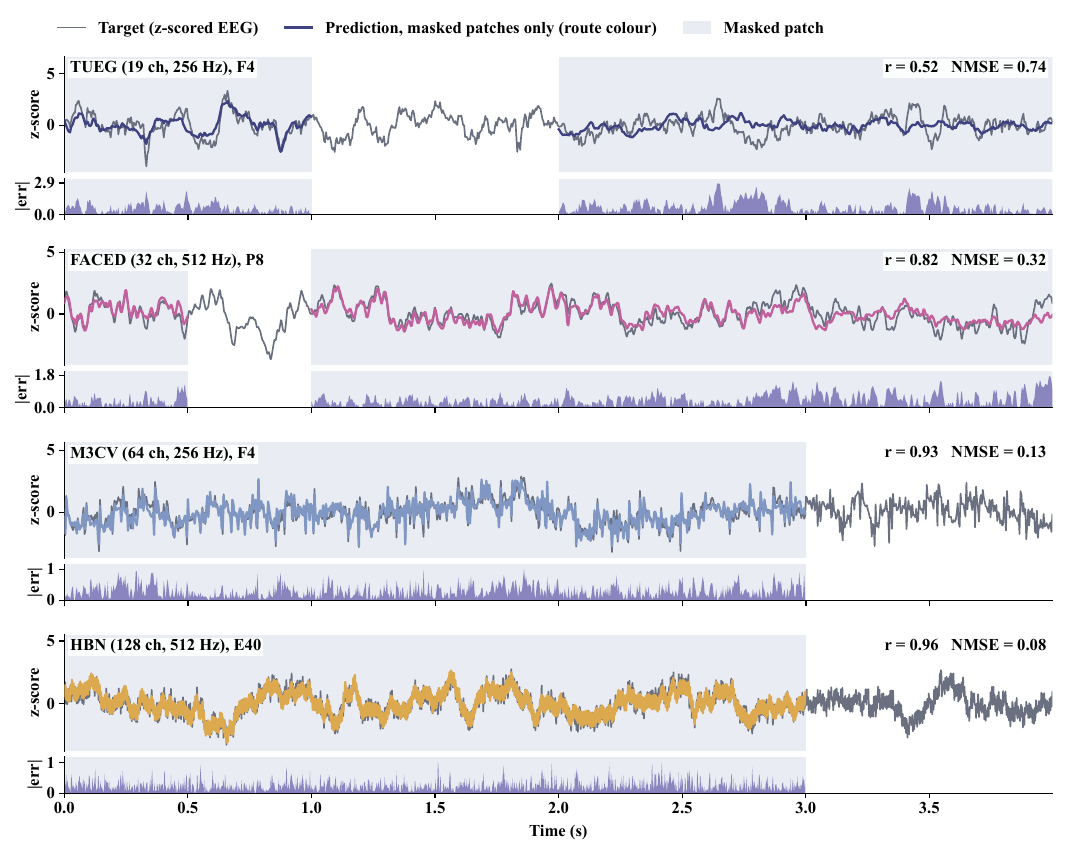}
  \caption{Single-channel views of the windows in
  Fig.~\ref{fig:mask_reconstruction}: F4, P8, F4, and E40, respectively.
  Targets are gray; predictions and absolute errors are shown on masked
  patches. Correlation and normalized mean squared error (NMSE) are evaluated
  on masked samples.}
  \label{fig:raw_waveform}
\end{figure}

Two shallow decoders per sampling rate then predict the masked patches in two
spaces. The spectral target is the clean front-end output, detached from the
gradient graph and normalized per patch; the raw target is the preprocessed
signal. Training minimizes
\begin{align}
  \mathcal{L} = \;&0.5\,\mathrm{MSE}\big(\hat{P}_{\mathrm{spec}},
  \mathrm{sg}[\tilde{P}]\big) \nonumber\\
  &+ 0.5\,\mathrm{SmoothL1}\big(\hat{P}_{\mathrm{raw}}, P\big) \nonumber\\
  &+ 10^{-3}\,\mathrm{KL}(\alpha\,\|\,u),
  \label{eq:pretrain}
\end{align}
where both reconstruction losses are evaluated on masked valid tokens,
$\mathrm{sg}[\cdot]$ denotes stop-gradient, and the last term regularizes the
folding weights in \eqref{eq:alpha}. The spectral target trains prediction in
the learned representation, the raw target anchors it to the measured signal,
and the stop-gradient prevents target-side updates within a training step.

Fig.~\ref{fig:mask_reconstruction} shows what the two heads recover on all four
EEG routes, from 19-channel TUEG at 256\,Hz to 128-channel HBN at 512\,Hz. For
each route, we select the validation window with median NMSE and mask 70\,\% of
its patches before the front end. Masked-patch correlations range from 0.79 to
0.89 for the raw-EEG head and from 0.75 to 0.88 for the folded-wavelet head,
with one shared backbone serving all four route-specific front ends.
Single-channel traces from the same windows (Fig.~\ref{fig:raw_waveform}) show
that predictions recover waveform structure across masked spans of up to three
consecutive seconds, reaching correlations of 0.93 and 0.96 on the selected
M3CV and HBN channels. The TUEG channel is harder ($r=0.52$), although its
window-level correlation is 0.79, indicating that reconstruction difficulty
varies across channels within one recording.

Each corpus is assigned to a compatible acquisition route rather than forced
into a single modality-wide layout. To prevent large datasets from dominating
pre-training, corpora are sampled with probability $P(d)\propto n_d^{0.5}$ for a
corpus of $n_d$ windows, while per-device batch sizes scale inversely with the
route token count. During preprocessing, recordings are standardized to common
physical units and, where necessary, resampled to the target route rate via
anti-aliased polyphase filtering. Channels are then mapped to their named slots
and each window is normalized, without applying any additional band-pass
filtering.

\subsection{Downstream Adaptation}
\label{sec:training:downstream}
A downstream task specifies its channel count, sampling rate, and sensor names.
The Transformer, channel embedding table, projection, and identity gate are
always transferred from the modality checkpoint. Front ends and patch
embeddings are transferred when the configuration matches a pre-training route,
as for PTB-XL on the 12-lead ECG route, and are newly initialized otherwise.
Mean-pooled tokens feed a two-layer prediction head, and we evaluate three
adaptation modes. \emph{Fine-tune} optimizes all parameters with layer-wise
learning-rate decay in the Transformer. \emph{Probe} freezes transferred
components and trains only the head and any newly introduced front-end or
adapter parameters. \emph{Scratch} trains the same architecture from random
initialization, isolating the contribution of pre-training under each task
protocol.

\subsection{Reliability-Aware Multimodal Fusion}
\label{sec:training:fusion}

\begin{figure*}[t]
  \centering
  \includegraphics[width=\textwidth]{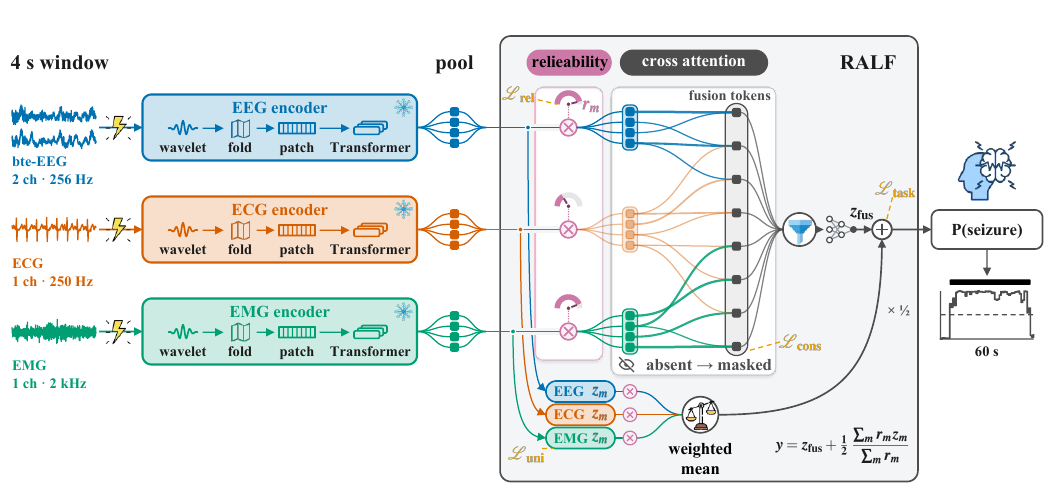}
  \caption{Reliability-aware latent fusion (RALF) of 4-s EEG, ECG, and EMG windows. Reliability scores gate tokens and weight unimodal predictions. Orange symbols denote training losses; the trace shows seizure probability over a 60-s test excerpt with a 0.5 threshold.}
  \label{fig:fusion}
\end{figure*}

PhysioWave combines unimodal logits with one learned weight per modality. These
weights are fixed after training, so they cannot reflect the degradation of a
particular sensor in a particular window. Reliability-aware latent fusion
(RALF; Fig.~\ref{fig:fusion}) fuses compact modality summaries and estimates
their reliability for every window. An attention pool with $q=4$ learnable
queries reduces each encoder's token sequence to four summary tokens, and
$n_f=8$ learnable fusion tokens attend to their concatenation through two
cross-attention layers before being pooled and classified. The fusion module
thus receives a fixed number of tokens per available modality, independent of
window length and channel count, although encoder and pooling costs still scale
with the input sequence.

For each modality, a sigmoid head predicts a reliability $r_m\in(0,1)$ from its
summary tokens. This score gates the modality's attention keys and values and
also weights a residual contribution from its unimodal logits:
\begin{equation}
  \mathbf{z} = \mathbf{z}_{\mathrm{fused}} +
  \frac{1}{2}\,\frac{\sum_{m\in\mathcal{M}}r_m\,\mathbf{z}_m}
  {\sum_{m\in\mathcal{M}}r_m},
  \label{eq:ralf}
\end{equation}
where $\mathcal{M}$ is the set of available modalities and $\mathbf{z}_m$ is the
output of a modality-specific prediction head. Normalization keeps the residual
scale consistent across modality subsets, and the denominator is clamped below
at $10^{-6}$ for numerical stability. Reliability therefore shapes both the
cross-modal interaction and the unimodal evidence. Because available subsets may
vary between items in a batch, absent modalities are excluded by attention masks
and omitted from \eqref{eq:ralf}; inputs with no available modality are
rejected.

During training, each modality is corrupted with probability $0.5$ before its
encoder by additive noise, baseline drift, saturation, bad channels, or temporal
dropout. A sampled severity $\sigma_m\in[0,1]$ defines the reliability target
$1-\sigma_m$, directly supervising quality estimation. Random modality dropping
further creates a subset view, whose prediction is encouraged to agree with the
full-set prediction. The objective is
\begin{align}
  \mathcal{L} = \;&\mathrm{CE}(\mathbf{z}, y)
  + 0.2\sum_{m\in\mathcal{M}}\mathrm{CE}(\mathbf{z}_m, y) \nonumber\\
  &+ \sum_{m\in\mathcal{M}}\big(r_m - (1-\sigma_m)\big)^2 \nonumber\\
  &+ \mathrm{KL}_{\mathrm{sym}}\big(\mathbf{z}^{\mathrm{subset}}\,\|\,\mathbf{z}\big),
  \label{eq:fusionloss}
\end{align}
where the symmetric KL term compares the class distributions of the two logit
vectors. The framework permits frozen, fine-tuned, or randomly initialized
encoders; the seizure-detection experiments use frozen encoders, connected to
the wearable acquisition configurations through task-specific front ends and
adapters.

\section{Experiments and Results}
\label{sec:experiments}
We first evaluate transfer across EEG, ECG, and sEMG tasks, then test whether
their representations can be combined for wearable seizure detection under
missing or corrupted inputs. EEG ablations isolate the contribution and
computational cost of the proposed components.

\begin{table}[hbp]
\centering
\caption{Pre-training corpora and input configurations.
The route $(C,f)$ gives channel count and sampling rate in Hz.
Counts refer to processed training windows.}
\label{tab:pretrain_sets}
\NSTTableStyle
\small
\renewcommand{\arraystretch}{1.12}
\setlength{\tabcolsep}{4pt}
\begin{tabular}{llrr}
\toprule
\rowcolor{NSTHead}
\textbf{Modality} & \textbf{Corpus}
& \textbf{Route} $(C,f)$ & \textbf{Windows} \\
\midrule

\rowcolor{NSTBaselineBlue}
 & TUEG~\cite{obeid2016temple}
 & (19, 256) & 13{,}710{,}516 \\
\rowcolor{NSTBaselineBlue}
 & HBN~\cite{alexander2017open}
 & (128, 512) & 1{,}352{,}818 \\
\rowcolor{NSTBaselineBlue}
 & FACED~\cite{chen2023large}
 & (32, 512) & 129{,}268 \\
\rowcolor{NSTBaselineBlue}
 & TDBRAIN~\cite{van2022two}
 & (32, 512) & 86{,}991 \\
\rowcolor{NSTBaselineBlue}
 & M3CV~\cite{huang2022m3cv}
 & (64, 256) & 81{,}895 \\
\rowcolor{NSTBaselineBlue}
 & PhysioNet MI~\cite{schalk2009eeg}
 & (64, 256) & 36{,}933 \\
\rowcolor{NSTBaselineBlue}
\multirow{-7}{*}{\textbf{EEG}} & HGD~\cite{schirrmeister2017deep}
 & (128, 512) & 23{,}074 \\
\midrule

\rowcolor{NSTBaselineLilac}
 & PulseDB~\cite{wang2023pulsedb}
 & (1, 250) & 4{,}992{,}774 \\
\rowcolor{NSTBaselineLilac}
 & Icentia11k~\cite{tan2019icentia11k}
 & (1, 250) & 1{,}650{,}343 \\
\rowcolor{NSTBaselineLilac}
 & MIMIC-IV-ECG~\cite{gow2023mimic}
 & (12, 500) & 748{,}592 \\
\rowcolor{NSTBaselineLilac}
 & CODE-15\%~\cite{ribeiro2021code}
 & (12, 500) & 123{,}202 \\
\rowcolor{NSTBaselineLilac}
 & SPH~\cite{liu2022large}
 & (12, 500) & 24{,}856 \\
\rowcolor{NSTBaselineLilac}
 & MedalCare-XL~\cite{gillette2023medalcare}
 & (12, 500) & 15{,}446 \\
\rowcolor{NSTBaselineLilac}
\multirow{-7}{*}{\textbf{ECG}} & Georgia~\cite{perez2020classification}
 & (12, 500) & 9{,}750 \\
\midrule

\rowcolor{NSTPeach!20!white}
 & emg2qwerty~\cite{sivakumar2024emg2qwerty}
 & (16, 2000) & 2{,}359{,}161 \\
\rowcolor{NSTPeach!20!white}
 & emg2pose~\cite{salter2024emg2pose}
 & (16, 2000) & 1{,}203{,}781 \\
\rowcolor{NSTPeach!20!white}
 & CEMHSEY $8{\times}8$~\cite{yang2025consecutive}
 & (64, 2000) & 662{,}710 \\
\rowcolor{NSTPeach!20!white}
 & CEMHSEY $5{\times}13$~\cite{yang2025consecutive}
 & (64, 2000) & 453{,}239 \\
\rowcolor{NSTPeach!20!white}
 & Hyser~\cite{jiang2021open}
 & (64, 2000) & 274{,}188 \\
\rowcolor{NSTPeach!20!white}
\multirow{-6}{*}{\textbf{sEMG}} & putEMG~\cite{kaczmarek2019putemg}
 & (24, 2000) & 153{,}124 \\
\bottomrule
\end{tabular}
\end{table}

\begin{table*}[!htbp]
\centering
\caption{EEG classification: balanced accuracy (BAC), Cohen's kappa,
weighted F1 (W-F1), and AUROC. Our values are three-seed means; evaluation
splits are specified in the text.
\NSTBest{Bold purple}: highest reported value; \NSTSecond{underline}: second highest.}
\label{tab:eeg_downstream}
\NSTTableStyle
\fontsize{9}{10.5}\selectfont
\setlength{\tabcolsep}{4pt}
\setlength{\NSTMetricWidth}{\dimexpr(\textwidth-120pt-31pt-88pt)/9\relax}
\begin{tabular}{V{120pt}N{31pt}*{9}{N{\NSTMetricWidth}}}
\toprule
\rowcolor{NSTHead}
 & & \multicolumn{3}{c}{\textbf{Sleep-EDFx}} &
 \multicolumn{3}{c}{\textbf{PhysioNet-P300}} &
 \multicolumn{3}{c}{\textbf{KaggleERN}} \\

\rowcolor{NSTHead}
\multirow{-2}{*}{\textbf{Method}} & \multirow{-2}{*}{\textbf{Params}} &
BAC & Kappa & W-F1 & BAC & Kappa & {\fontsize{8}{9}\selectfont AUROC} & BAC & Kappa & {\fontsize{8}{9}\selectfont AUROC} \\
\midrule
\rowcolor{NSTBaselineBlue}
BENDR \NSTVenue{Front.\ Hum.\ Neurosci.'21} & 157M &
0.6655 & 0.6659 & 0.7507 & 0.6114 & 0.2227 & 0.6588 & 0.5672 & 0.1461 & 0.6030 \\
\rowcolor{NSTBaselineBlue}
BIOT\NSTVenue{NeurIPS'23} & 3.2M &
0.6622 & 0.6461 & 0.7415 & 0.5485 & 0.0968 & 0.5308 & 0.5118 & 0.0297 & 0.5495 \\
\rowcolor{NSTBaselineBlue}
LaBraM\NSTVenue{ICLR'24} & 5.8M &
0.6771 & 0.6710 & 0.7592 & 0.6477 & 0.2935 & 0.7068 & 0.5439 & 0.0944 & 0.5693 \\
\rowcolor{NSTBaselineBlue}
EEGPT \NSTVenue{NeurIPS'24} & 25M &
0.6917 & 0.6857 & 0.7654 & 0.6502 & \NSTBest{0.2999} & 0.7168 & \NSTSecond{0.5837} & \NSTSecond{0.1882} & \NSTSecond{0.6621} \\
\midrule
\rowcolor{NSTOurs}
\method{} (scratch) & 30M &
\NSTSecond{0.6924} & \NSTSecond{0.7363} & \NSTSecond{0.8043} & 0.6512 & 0.2337 & 0.7194 & 0.5704 & 0.1409 & 0.6001 \\
\rowcolor{NSTOurs}
\method{} (probe) & 30M &
0.6654 & 0.7200 & 0.7897 & \NSTSecond{0.6612} & 0.2665 & \NSTSecond{0.7266} & 0.5369 & 0.0670 & 0.5703 \\
\rowcolor{NSTFocus}
\textbf{\method{} (fine-tune)} & 30M &
\NSTBest{0.7060} & \NSTBest{0.7365} & \NSTBest{0.8092} & \NSTBest{0.6706} & \NSTSecond{0.2975} & \NSTBest{0.7370} & \NSTBest{0.6171} & \NSTBest{0.2059} & \NSTBest{0.6632} \\
\bottomrule
\end{tabular}
\end{table*}

\subsection{Datasets and Training Settings}
\label{sec:settings}

\textbf{Pre-training data.}
Each recording is assigned to a front-end route according to its channel count
and sampling rate. Table~\ref{tab:pretrain_sets} summarizes the input
configurations and available window counts. EEG pre-training uses 15.4\,M
four-second windows from clinical recordings in TUEG~\cite{obeid2016temple},
research recordings in TDBRAIN~\cite{van2022two} and FACED~\cite{chen2023large},
BCI recordings in M3CV~\cite{huang2022m3cv} and PhysioNet MI~\cite{schalk2009eeg},
and high-density recordings in HBN~\cite{alexander2017open} and
HGD~\cite{schirrmeister2017deep}.

ECG pre-training uses 7.6\,M ten-second windows spanning 12-lead recordings
from MIMIC-IV-ECG~\cite{gow2023mimic}, CODE-15\%~\cite{ribeiro2021code},
SPH~\cite{liu2022large}, and Georgia~\cite{perez2020classification},
synthetic ECGs from MedalCare-XL~\cite{gillette2023medalcare}, and
single-lead recordings from PulseDB~\cite{wang2023pulsedb} and
Icentia11k~\cite{tan2019icentia11k}. The sEMG encoder uses 5.1\,M one-second
windows from 16-electrode wristbands
(emg2qwerty~\cite{sivakumar2024emg2qwerty} and
emg2pose~\cite{salter2024emg2pose}), 24-channel forearm recordings
(putEMG~\cite{kaczmarek2019putemg}), and 64-channel inputs extracted from
high-density arrays (CEMHSEY~\cite{yang2025consecutive} and
Hyser~\cite{jiang2021open}). Compared with PhysioWave, these corpora add
single-lead ECG and several sEMG configurations, while EEG pre-training is new.

\textbf{Optimization.}
All encoders use the objective in Section~\ref{sec:training:pretrain}, a
70\,\% mask ratio, and AdamW with $\beta_1=0.9$, $\beta_2=0.98$, and
weight decay $0.01$. A linear warm-up precedes cosine learning-rate decay,
and training uses mixed precision. Corpora are sampled with temperature
$0.5$; route-specific batch sizes give approximately equal token counts per
step. EEG and ECG encoders are trained for 60 epochs and the sEMG encoder
for 40 epochs on NVIDIA A100 GPUs. Peak device memory is 26\,GB for
128-channel EEG, 19\,GB for ECG, and 16\,GB for 64-channel sEMG under these
training configurations. Learning rates and warm-up lengths are set per
modality, and the complete hyperparameters are provided in the released
configuration files.

\textbf{Evaluation and statistical reporting.}
We evaluate three adaptation modes defined in
Section~\ref{sec:training:downstream}: fine-tuning all parameters with
layer-wise learning-rate decay, probing with transferred components frozen,
and training the same architecture from scratch. Hyperparameters and
checkpoints are selected using validation data, with balanced accuracy or
Cohen's kappa for multi-class tasks and AUROC for binary and multi-label
tasks. All performance results from experiments conducted in this study,
including ablations and rerun baselines, are arithmetic means over three
independent random seeds.
Values quoted from previous work retain the original studies' aggregation
and evaluation protocols. Benchmarks use subject-disjoint splits unless a
task-specific protocol is stated below. For multi-label ECG tasks,
decision thresholds are selected on validation data and then fixed for
record-level test evaluation.

\subsection{EEG Results}
\label{sec:results:eeg}
We evaluate three tasks following the exact evaluation protocol of
EEGPT~\cite{wang2024eegpt}: five-stage sleep staging on
Sleep-EDFx~\cite{kemp2000analysis}, target detection on
PhysioNet-P300~\cite{citi2010documenting}, and error-related negativity
detection on KaggleERN~\cite{margaux2012objective}. Table~\ref{tab:eeg_downstream}
compares \method with BENDR~\cite{kostas2021bendr}, BIOT~\cite{yang2023biot},
LaBraM~\cite{jiang2024large}, and EEGPT under the same subject-disjoint
cross-validation folds, with baseline results cited directly from~\cite{wang2024eegpt}.

Fine-tuned \method achieves the highest value on eight of nine metrics, and the
margins are largest where the baselines are weakest. On Sleep-EDFx, its Fpz-Cz
and Pz-Oz derivations at 100\,Hz match no pre-training route or vocabulary
entry, so the front end and both channel embeddings are learned downstream.
Even so, \method raises kappa from 0.686 to 0.737 and weighted F1 from 76.5\,\%
to 80.9\,\% over EEGPT. The scratch model alone already reaches a kappa of
0.736, which indicates that the multiscale front end itself suits sleep
staging, a task defined by band-limited rhythms; Section~\ref{sec:analysis:fold}
shows the fold shifting weight toward the slowest scale in deep sleep. On
KaggleERN, \method improves balanced accuracy by 3.3 points and kappa by 0.018
over EEGPT, the largest balanced-accuracy gain among the three tasks. On
PhysioNet-P300, balanced accuracy and AUROC each rise by 2.0 points, while kappa
is within 0.003 of EEGPT.

The adaptation modes show that tasks draw on pre-training in different ways.
On PhysioNet-P300, the frozen encoder already exceeds every baseline in
balanced accuracy and AUROC, and fine-tuning adds only 0.9 points, so the
pre-trained representation captures the P300 response largely as is. KaggleERN
behaves differently: probing falls 3.4 points below scratch, yet fine-tuning
exceeds scratch by 4.7 balanced-accuracy and 6.3 AUROC points. The pre-trained
weights therefore contain information relevant to error-related negativity, but
backbone adaptation is needed to expose it. On Sleep-EDFx, pre-training adds
1.4 balanced-accuracy points over scratch with nearly unchanged kappa,
consistent with the strong contribution of our architecture.

\begin{table}[htbp]
\centering
\caption{ECG classification: macro-F1 and macro-AUROC in \%.
Our results are means over three seeds; baselines are quoted from PhysioWave.
\NSTBest{Bold purple}: highest reported value; \NSTSecond{underline}: second highest.}
\label{tab:ecg_downstream}
\NSTTableStyle
\fontsize{7.2}{8.4}\selectfont
\setlength{\tabcolsep}{1.5pt}
\renewcommand{\arraystretch}{1.18}
\setlength{\NSTMetricWidth}{\dimexpr(\columnwidth-74pt-24pt-24pt)/6\relax}
\begin{tabular}{V{74pt}N{24pt}*{6}{N{\NSTMetricWidth}}}
\toprule
\rowcolor{NSTHead}
 & & \multicolumn{2}{c}{\textbf{PTB-XL Super}} &
 \multicolumn{2}{c}{\textbf{CPSC 2018}} &
 \multicolumn{2}{c}{\textbf{Chapman}} \\

\rowcolor{NSTHead}
\multirow{-2}{*}{\textbf{Method}} & \multirow{-2}{*}{{\fontsize{6.5}{7.5}\selectfont\textbf{Params}}} &
F1 & {\fontsize{6.5}{7.5}\selectfont AUC} & F1 & {\fontsize{6.5}{7.5}\selectfont AUC} & F1 & {\fontsize{6.5}{7.5}\selectfont AUC} \\
\midrule
\rowcolor{NSTBaselineBlue}
MERL \NSTVenue{ICML'24} & 11M & 48.1 & 91.9 & 72.8 & 92.6 & --- & 87.9 \\
\rowcolor{NSTBaselineBlue}
MaeFE \NSTVenue{TIM'23} & 9M & 64.7 & 88.6 & 71.6 & 94.5 & --- & --- \\
\rowcolor{NSTBaselineBlue}
OpenECG-SimCLR \NSTVenue{arXiv'25} & 11M & 46.9 & 91.5 & \NSTSecond{73.1} & 92.4 & 52.3 & 95.1 \\
\rowcolor{NSTBaselineBlue}
OpenECG-BYOL \NSTVenue{arXiv'25} & 11M & 47.7 & 91.1 & 72.8 & 92.6 & 51.5 & 94.8 \\
\rowcolor{NSTBaselineBlue}
OpenECG-MAE \NSTVenue{arXiv'25} & 11M & 48.1 & 90.9 & \NSTBest{74.5} & 93.2 & 50.8 & 94.2 \\
\rowcolor{NSTBaselineLilac}
PhysioWave-S \NSTVenue{NeurIPS'25} & 5M & 65.8 & 92.7 & 71.6 & 95.5 & 52.1 & 96.4 \\
\rowcolor{NSTBaselineLilac}
PhysioWave-B \NSTVenue{NeurIPS'25} & 15M & 64.5 & \NSTSecond{93.4} & 72.5 & \NSTSecond{95.9} & 53.8 & 97.2 \\
\rowcolor{NSTBaselineLilac}
PhysioWave-L \NSTVenue{NeurIPS'25} & 37M & 66.7 & \NSTBest{94.6} & \NSTSecond{73.1} & \NSTBest{96.1} & 54.8 & 98.3 \\
\midrule
\rowcolor{NSTOurs}
\method{} (scratch) & 30M & \NSTSecond{66.9} & 87.7 & 64.4 & 92.5 & \NSTSecond{55.8} & \NSTSecond{98.5} \\
\rowcolor{NSTOurs}
\method{} (probe) & 30M & 64.6 & 85.6 & 58.7 & 89.5 & 53.4 & 95.7 \\
\rowcolor{NSTFocus}
\textbf{\method{} (fine-tune)} & 30M & \NSTBest{70.9} & 90.3 & \NSTSecond{73.1} & 94.2 & \NSTBest{56.7} & \NSTBest{99.4} \\
\bottomrule
\end{tabular}
\end{table}

\subsection{ECG Results}
\label{sec:results:ecg}
ECG evaluation covers the five diagnostic superclasses of
PTB-XL~\cite{wagner2020ptb}, the nine classes of
CPSC~2018~\cite{liu2018open}, and Chapman--Shaoxing~\cite{zheng202012}.
All three datasets consist of 12-lead recordings at 500\,Hz, enabling direct
transfer of the corresponding front end. Following the exact data splits and
evaluation protocol of PhysioWave~\cite{chen2026physiowave},
Table~\ref{tab:ecg_downstream} reports macro-F1 and macro-AUROC alongside the
baseline results cited from PhysioWave.

\method achieves the highest macro-F1 on PTB-XL Super and Chapman--Shaoxing,
70.9\,\% and 56.7\,\%, with fewer parameters than PhysioWave-L (30\,M versus
37\,M). On PTB-XL Super, the margin is 4.2 points over PhysioWave-L and 6.2
points over MaeFE, and Chapman--Shaoxing also yields the highest macro-AUROC,
99.4\,\%. On CPSC~2018, macro-F1 ties PhysioWave-L and OpenECG-SimCLR at
73.1\,\%, 1.4 points below OpenECG-MAE. The two metrics diverge on PTB-XL Super
and CPSC~2018, where PhysioWave-L keeps the higher macro-AUROC. Macro-F1 is
measured at the operating threshold and weights every class equally, whereas
macro-AUROC summarizes ranking across all thresholds. The pattern therefore
indicates that \method makes more balanced decisions across diagnostic classes,
while PhysioWave-L ranks records slightly better on these two datasets. In the
five-class, single-label PTB-XL task, the same encoder reaches 73.9\,\%
accuracy.

Pre-training improves every ECG metric over scratch, with the largest benefit on
CPSC~2018 and the smallest on Chapman--Shaoxing. CPSC~2018 gains 8.7 macro-F1
and 1.7 AUROC points, PTB-XL Super gains 4.0 and 2.6, and Chapman--Shaoxing,
where scratch already reaches 98.5\,\% AUROC, gains 0.9 on both. Five-class
PTB-XL accuracy rises by 3.4 points. The frozen encoder retains 95--96\,\% of
fine-tuned AUROC on all three multi-label benchmarks, but its share of
fine-tuned macro-F1 varies more widely: about 80\,\% on CPSC~2018, compared with
91\,\% on PTB-XL Super and 94\,\% on Chapman--Shaoxing. The pre-trained
representation thus already orders records by diagnosis, and fine-tuning mainly
improves the threshold-level decisions captured by macro-F1, most visibly on the
nine-class CPSC~2018 task.

\subsection{sEMG Results}
\label{sec:results:emg}
The sEMG benchmarks test transfer to two configurations absent from
pre-training: the eight-electrode, 200\,Hz EPN-612
wristband~\cite{benalcazar2020emg} and the 28-electrode, 2\,kHz GRABMyo
array~\cite{pradhan2022multi}. Both receive a new front end, while the
pre-trained backbone and channel vocabulary are retained. EPN-612 follows
PhysioWave's user-dependent protocol, with baseline values quoted from that
paper~\cite{chen2026physiowave}. GRABMyo trains on sessions 1--2 and tests on
session 3, and all baselines are rerun on this split with three seeds. As
Table~\ref{tab:emg_downstream} shows, \method achieves the highest accuracy and
macro-F1 on both benchmarks with 12\,M parameters, about one third of
PhysioWave-L. On EPN-612, its 94.80\,\% accuracy exceeds PhysioWave-L by 0.30
points; on GRABMyo, accuracy and macro-F1 reach 89.78\,\% and 89.74\,\%, 1.25
and 1.37 points above PhysioWave-L. The generic time-series models trail by a
wider margin on GRABMyo, where \method exceeds MOMENT by 7.11 points with
roughly one thirtieth of its parameters and OTiS~\cite{turgut2024towards} by
10.40 points.

The two protocols reveal different roles for pre-training. Under EPN-612's
user-dependent protocol, scratch training already reaches 94.34\,\%, and
fine-tuning adds only 0.46 points, since training and test data come from the
same users. GRABMyo instead tests generalization across recording sessions,
where electrode re-placement and changing skin contact shift the signal
distribution. Here scratch drops to 85.16\,\%, while the frozen encoder reaches
87.41\,\%, matching PhysioWave-B, and fine-tuning adds a further 2.37 points.
Pre-training thus contributes most where the downstream data alone cannot cover
session variability. Probing is less effective on EPN-612, falling 4.45 points
below scratch. This gap coincides with a spectral mismatch: the 200\,Hz
wristband retains content only below 100\,Hz, whereas the pre-trained sEMG
front ends emphasize roughly 100 to 200\,Hz (Section~\ref{sec:analysis:wavelet}),
so the frozen backbone receives inputs outside its pre-training range until
fine-tuning adapts it.

\begin{table}[t]
\centering
\caption{sEMG gesture recognition: accuracy and macro-F1 in \%.
Our runs are means over three seeds; EPN-612 baselines retain their published values.
\NSTBest{Bold purple}: highest reported value; \NSTSecond{underline}: second highest.}
\label{tab:emg_downstream}
\NSTTableStyle
\fontsize{8}{9.5}\selectfont
\setlength{\tabcolsep}{2.8pt}
\setlength{\NSTMetricWidth}{\dimexpr(\columnwidth-97pt-28pt-33.6pt)/4\relax}
\begin{tabular}{V{97pt}N{28pt}*{4}{N{\NSTMetricWidth}}}
\toprule
\rowcolor{NSTHead}
 & & \multicolumn{2}{c}{\textbf{EPN-612}} & \multicolumn{2}{c}{\textbf{GRABMyo}} \\

\rowcolor{NSTHead}
\multirow{-2}{*}{\textbf{Method}} & \multirow{-2}{*}{\textbf{Params}} &
Acc. & F1 & Acc. & F1 \\
\midrule
\rowcolor{NSTBaselineBlue}
MOMENT \NSTVenue{ICML'24} & 385M & 90.87 & 90.16 & 82.67 & 82.19 \\
\rowcolor{NSTBaselineBlue}
OTiS \NSTVenue{arXiv'24} & 45M & 87.55 & 88.03 & 79.38 & 78.80 \\
\rowcolor{NSTBaselineLilac}
PhysioWave-S \NSTVenue{NeurIPS'25} & 5M & 93.12 & 93.40 & 85.26 & 85.03 \\
\rowcolor{NSTBaselineLilac}
PhysioWave-B \NSTVenue{NeurIPS'25} & 15M & 93.68 & 93.91 & 87.40 & 87.16 \\
\rowcolor{NSTBaselineLilac}
PhysioWave-L \NSTVenue{NeurIPS'25} & 37M & \NSTSecond{94.50} & \NSTSecond{94.56} & \NSTSecond{88.53} & \NSTSecond{88.37} \\
\midrule
\rowcolor{NSTOurs}
\method{} (scratch) & 12M & 94.34 & 94.34 & 85.16 & 85.13 \\
\rowcolor{NSTOurs}
\method{} (probe) & 12M & 89.89 & 89.88 & 87.41 & 87.44 \\
\rowcolor{NSTFocus}
\textbf{\method{} (fine-tune)} & 12M & \NSTBest{94.80} & \NSTBest{94.79} & \NSTBest{89.78} & \NSTBest{89.74} \\
\bottomrule
\end{tabular}
\end{table}

\subsection{Multimodal Fusion}
\label{sec:results:fusion}

\begin{figure*}[t]
  \centering
  \includegraphics[width=\textwidth]{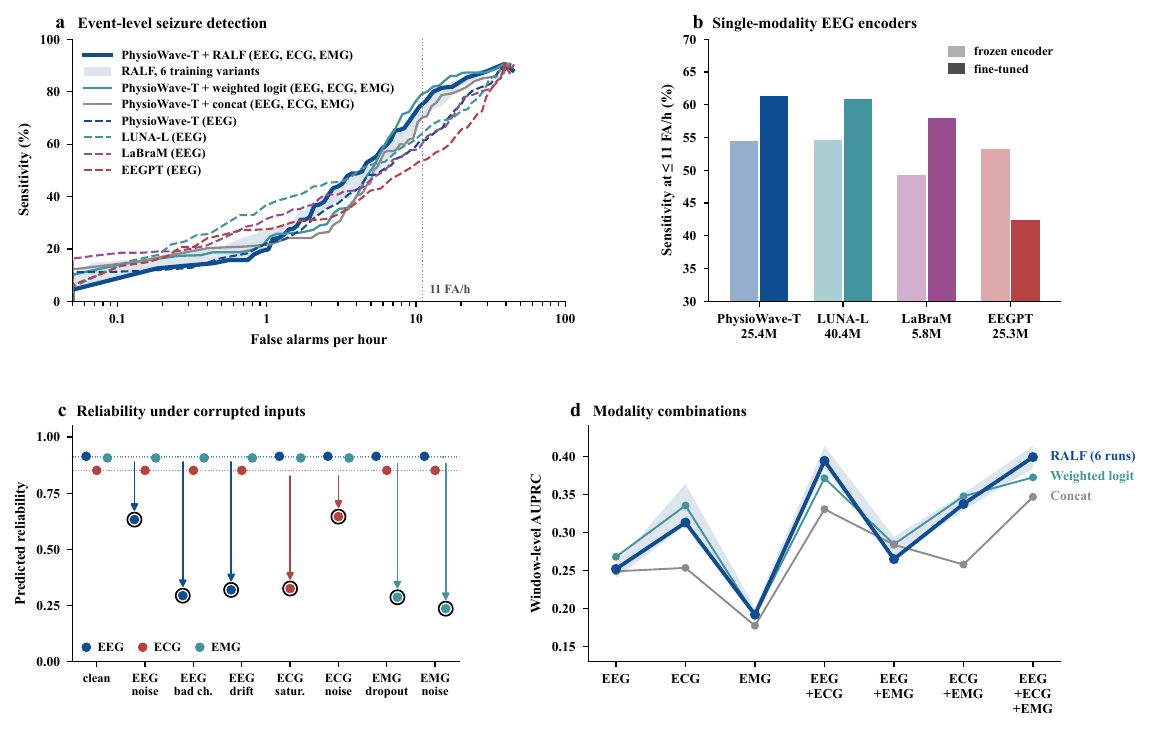}
  \caption{Seizure detection on SeizeIT2 (official split, 29 test patients,
  2{,}912\,h, 181 seizures). (a)~Event-level sensitivity against false alarms
  per hour for the three fusion rules over the frozen \method encoders and for
  each EEG-only encoder at its better of frozen and fine-tuned; the dotted line
  is the 11 false alarms per hour of the dataset paper's SVM. (b)~Frozen and
  fine-tuned EEG encoders at $\leq$11 false alarms per hour, with encoder size.
  (c)~RALF's predicted reliability on clean inputs and with one modality
  corrupted at severity 0.7; the ringed point is the corrupted modality.
  (d)~Window-level AUPRC of every modality subset. The RALF line is the
  original training configuration; the shaded band in (a) and (d) is the
  min--max over six RALF training variants, the original and five
  learning-rate and schedule settings.}
  \label{fig:fusion_results}
\end{figure*}

SeizeIT2~\cite{bhagubai2025seizeit2} records behind-the-ear EEG, ECG, and EMG
simultaneously and tests whether the independently pre-trained encoders can be
combined for wearable seizure monitoring. RALF fuses the three frozen \method
encoders and is compared with concatenation and PhysioWave's weighted-logit
fusion over the same encoders, while LUNA~\cite{doner2026luna},
LaBraM~\cite{jiang2024large}, and EEGPT~\cite{wang2024eegpt} serve as
single-modality EEG references. We report window-level AUPRC and event-level
sensitivity against false alarms per hour, and remove modalities or corrupt one
at a time with noise, bad channels, baseline drift, saturation, or dropout at
severity 0.7 (Fig.~\ref{fig:fusion_results}). The EEG branch alone is a
demanding transfer target: its two behind-the-ear channels match no
pre-training route and receive a new front end. Fine-tuned, \method reaches
61.4\,\% sensitivity at $\leq$11 false alarms per hour, ahead of LUNA-L
(60.9\,\%), LaBraM (58.0\,\%), and EEGPT (42.4\,\%), with 25.4\,M parameters
against LUNA-L's 40.4\,M. Frozen, \method and LUNA-L are nearly tied at
54.5\,\% and 54.6\,\%, and EEGPT is the only encoder that loses sensitivity when
fine-tuned (Fig.~\ref{fig:fusion_results}b), so adaptation to so sparse a
montage depends on a backbone that tolerates it.

Adding ECG and EMG changes the picture substantially. With frozen encoders,
fusion lifts event sensitivity at 11 false alarms per hour---the operating point
of the dataset paper's SVM baseline---from 61.4\,\% for fine-tuned \method alone
to 74.4\,\% with RALF and 78.9\,\% with weighted-logit fusion, exceeding any
EEG-only encoder by at least 13 points. Seizures therefore leave complementary
signatures across the three modalities, consistent with the ictal cardiac and
motor changes accompanying many events. Yet the fusion rule governs how
effectively this evidence is exploited: on window-level AUPRC (the validation
selection metric), RALF reaches 0.399 against 0.372 for weighted-logit fusion
and 0.347 for concatenation, and all six RALF training variants outperform both
whenever EEG and ECG are jointly present (0.384--0.414 on EEG+ECG and the
complete set). Sample-level reliability estimation also pays off under strict
alarm budgets, the regime that matters most for ambulatory monitoring: at 2
false alarms per hour, RALF detects 32.0\,\% of seizures, versus 26.7\,\% for
weighted-logit fusion.

The reliability scores explain this behavior. On clean inputs, RALF assigns
0.92, 0.85, and 0.91 to EEG, ECG, and EMG, and corrupting one modality lowers
only its own score, to 0.24--0.65, as expected from modality-local reliability
heads. The drop tracks how much signal survives: bad channels, drift,
saturation, dropout, and EMG noise, which swamps EMG's broadband content, push
reliability to 0.24--0.33, whereas noise on EEG or ECG leaves their dominant
rhythms visible and lowers it only to 0.63--0.65. Predictions degrade
correspondingly little. ECG noise costs 0.004 AUPRC, EMG dropout or noise
leaves it essentially at the clean value of 0.399, so RALF does not rely on EMG
when that signal is compromised, and AUPRC stays at or above 0.33 under every
tested corruption.

\subsection{Ablation Study}
\label{sec:ablation}
Table~\ref{tab:eeg_ablation} tests six aspects of the EEG encoder on
PhysioNet-P300 and KaggleERN. All values are means over three seeds.
The full, randomly initialized, and frozen models correspond to the
fine-tune, scratch, and probe settings in Table~\ref{tab:eeg_downstream}.
Each architectural ablation changes one component while retaining the
other settings and validation procedure. Label-efficiency comparisons
vary supervision, with a matched scratch condition at 10\,\% labels.
Electrode dropout is applied only at test time, with affected channels
zeroed and marked absent. $\Delta$ denotes the average change across
the four metrics relative to the full model, in percentage points.

\begin{table}[t]
\centering

\caption{EEG ablations on PhysioNet-P300 and KaggleERN, averaged over three seeds.
Metrics are percentages. $\Delta$ is the mean change across the four metrics
relative to the full model; darker shading indicates a larger decrease.}
\label{tab:eeg_ablation}

\NSTTableStyle
\renewcommand{\arraystretch}{1.12}
\fontsize{7.4}{8.6}\selectfont
\setlength{\tabcolsep}{2.2pt}
\setlength{\NSTMetricWidth}{%
  \dimexpr(\columnwidth-104pt-26.4pt)/5\relax
}

\begin{tabular}{V{104pt}*{5}{N{\NSTMetricWidth}}}
\toprule

\rowcolor{NSTHead}
&
\multicolumn{2}{c}{\textbf{P300}} &
\multicolumn{2}{c}{\textbf{KaggleERN}} &
\\

\rowcolor{NSTHead}
\multirow{-2}{*}{\textbf{Variant}} &
BAC &
{\fontsize{6.5}{7.5}\selectfont AUROC} &
BAC &
{\fontsize{6.5}{7.5}\selectfont AUROC} &
\multirow{-2}{*}{$\boldsymbol{\Delta}$} \\

\midrule

\rowcolor{NSTFocus}
\textbf{\method{} (full)}
& \textbf{67.1}
& \textbf{73.7}
& \textbf{61.7}
& \textbf{66.3}
& --- \\

\midrule

\NSTGroupRow{Scale folding (vs.\ dynamic fold)}

\NSTIndent Scale expansion
& 66.8 & 73.4 & 61.4 & 66.0 & \NSTDropA{0.3} \\

\NSTIndent Static uniform fold ($\alpha{=}u$)
& 65.6 & 72.0 & 60.2 & 64.5 & \NSTDropB{1.6} \\

\NSTIndent w/o synthesis filters ($g_s{=}\delta$)
& 66.4 & 73.0 & 61.0 & 65.6 & \NSTDropA{0.7} \\

\NSTGroupRow{Pre-training and adaptation}

\NSTIndent w/o pre-training
& 65.1 & 71.9 & 57.0 & 60.0 & \NSTDropC{3.7} \\

\NSTIndent Frozen encoder
& 66.1 & 72.7 & 53.7 & 57.0 & \NSTDropC{4.8} \\

\NSTIndent Tune top-2 blocks
& 66.0 & 72.4 & 57.9 & 61.6 & \NSTDropB{2.7} \\

\NSTIndent Tune top-4 blocks
& 65.9 & 72.2 & 59.8 & 64.1 & \NSTDropB{1.7} \\

\NSTGroupRow{Channel identity}

\NSTIndent w/o channel enc.\ ($g{=}0$)
& 64.5 & 70.9 & 60.8 & 65.2 & \NSTDropB{1.8} \\

\NSTIndent Shuffled channel names
& 64.1 & 70.5 & 60.6 & 65.0 & \NSTDropB{2.1} \\

\NSTGroupRow{Readout (vs.\ mean pooling)}

\NSTIndent Temporal head~\cite{wang2024eegpt}
& 63.4 & 69.9 & 60.9 & 65.4 & \NSTDropB{2.3} \\

\NSTIndent Attentive stat.\ pooling
& 65.4 & 71.6 & 61.3 & 66.0 & \NSTDropA{1.1} \\

\NSTGroupRow{Label efficiency}

\NSTIndent 25\,\% labels
& 64.2 & 70.1 & 59.6 & 63.8 & \NSTDropB{2.8} \\

\NSTIndent 10\,\% labels
& 62.6 & 68.0 & 57.8 & 61.2 & \NSTDropC{4.8} \\

\NSTIndent 10\,\% labels, w/o pre-train.
& 60.4 & 65.2 & 53.1 & 55.4 & \NSTDropD{8.7} \\

\NSTGroupRow{Missing electrodes}

\NSTIndent 25\,\% dropped
& 64.6 & 70.6 & 60.4 & 64.7 & \NSTDropB{2.1} \\

\NSTIndent 50\,\% dropped
& 62.9 & 68.6 & 58.3 & 62.2 & \NSTDropC{4.2} \\

\bottomrule
\end{tabular}
\end{table}

\textbf{Scale folding.}
Scale expansion feeds all $J{+}1$ bands to the backbone as separate token
rows, giving the Transformer direct access to every subband at four times the
sequence length. The dynamic fold scores 0.3 points higher on average with a
quarter of the tokens, so compressing the scales before tokenization loses
nothing that the backbone could have recovered from the longer sequence. How
the scales are combined does matter, however. A static uniform fold drops the
average by 1.6 points and falls 1.3 points below scale expansion, so a plain
average of the subbands discards information that the expanded representation
preserves. Of the 1.6-point gap between the uniform and dynamic folds, the
learned synthesis filters account for 0.7 points, since replacing every $g_s$
with a delta costs that much, and the data-dependent weights $\alpha$ account
for the remaining 0.9. Both components are therefore needed to match and then
exceed scale expansion, and the full fold achieves this with a predictor
shared across channels.

\begin{figure}[tbp]
  \centering
  \includegraphics[width=\columnwidth]{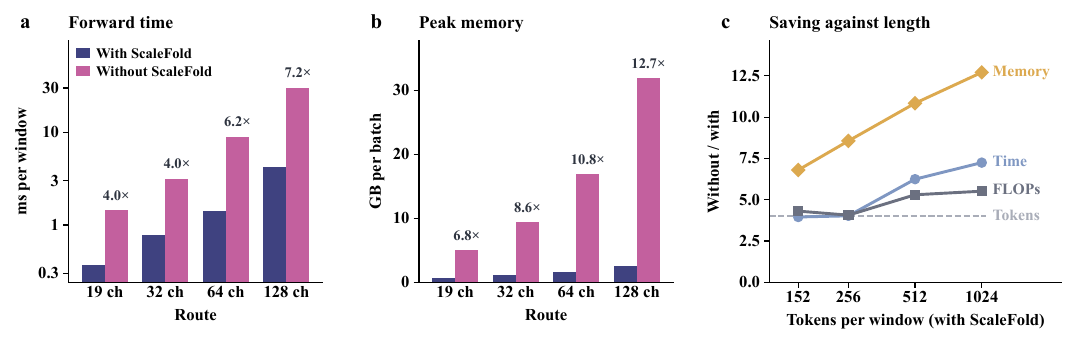}
  \caption{EEG encoder efficiency on one A100 in bf16, averaged over three seeds. (a)~Latency per window. (b)~Peak memory per batch. (c)~Savings versus token count.}
  \label{fig:fold_efficiency}
\end{figure}

Fig.~\ref{fig:fold_efficiency} quantifies the cost side on one A100 in bf16
with matched route-specific batch sizes. Across the four EEG routes, folding
reduces Transformer latency by 4.0--7.2$\times$, peak memory by
6.8--12.7$\times$, and FLOPs by 4.3--5.5$\times$. Memory shrinks more than
FLOPs because the attention maps, whose size grows quadratically with sequence
length, dominate activation memory; latency shrinks less than memory because
the front end and the linear projections keep their cost. All three savings
grow with token count and peak on the 128-channel route, where a batch needs
about 32\,GB without folding and 2.5\,GB with it in this profiling setup.
These figures use matched batch sizes and therefore differ from the
pre-training memory reported in Section~\ref{sec:settings}. The fold itself,
including its subband statistics and weighting network, accounts for only
1.5--2.6\,\% of the forward pass, so its accuracy benefit comes at negligible
cost.

\textbf{Pre-training and adaptation.}
Removing pre-training lowers the four-metric average by 3.7 points, with the
loss concentrated on KaggleERN, where balanced accuracy and AUROC fall by 4.7
and 6.3 points. The adaptation ladder shows why the two tasks differ. On
PhysioNet-P300, the frozen encoder is within one balanced-accuracy point of
full fine-tuning, and tuning the top two or four blocks changes little, so the
pre-trained representation already separates target from non-target responses.
On KaggleERN, the frozen encoder reaches only 53.7\,\%, below the 57.0\,\% of
the scratch model, and balanced accuracy climbs monotonically with the number
of tuned blocks: 57.9\,\% with two, 59.8\,\% with four, and 61.7\,\% with the
full backbone. Error-related negativity is a brief, low-amplitude deflection
embedded in ongoing frontocentral activity, and pre-training on four-second
windows of clinical EEG does not isolate it on its own. The pre-trained weights
nonetheless provide the better starting point, since fine-tuning from them ends
4.7 points above training from scratch.

\textbf{Channel identity.}
Setting the channel-encoding gate to zero in \eqref{eq:channel} lowers the
average score by 1.8 points, and shuffling channel names lowers it by 2.1. The
second number is the more informative one: the shuffled model keeps the same
embedding parameters and receives spatial information that is simply wrong,
and it performs worse than a model given no spatial information at all. The
benefit of correct identity therefore comes from the scalp geometry the
embeddings encode (Section~\ref{sec:analysis:topology}), not from added
capacity. The losses concentrate on PhysioNet-P300, where removing channel
encoding costs 2.6 balanced-accuracy and 2.8 AUROC points, compared with 0.9
and 1.1 on KaggleERN. The P300 response has a well-defined parietal
distribution, so knowing which electrode produced each token helps locate it;
the error-related negativity is spatially more diffuse, and the task relies
less on electrode identity.

\textbf{Readout.}
Mean pooling outperforms the EEGPT temporal head by 2.3 points and attentive
statistics pooling by 1.1 points on average, with the temporal head losing 3.7
balanced-accuracy points on PhysioNet-P300. Both alternatives add trainable
readout capacity on top of the backbone, and both hurt. Under full fine-tuning,
the backbone can already arrange its tokens so that a simple average is
linearly separable, and the extra readout parameters mainly add variance on
datasets of this size. The temporal head, designed for probing a frozen
encoder in EEGPT, is the more complex of the two and suffers most.

\textbf{Label efficiency.}
Reducing supervision to 10\,\% of the labels costs the pre-trained encoder 4.8
points on average, against 8.7 points for the matched scratch model, so
pre-training roughly halves the loss from scarce labels. The absolute numbers
on KaggleERN are more telling: with 10\,\% of the labels, the pre-trained
encoder already exceeds the scratch model trained on all labels, by 0.8
balanced-accuracy and 1.2 AUROC points, and with 25\,\% the margins widen to
2.6 and 3.8 points. One tenth of the annotations is therefore enough to match
full-data training from scratch on this task. The effect is weaker on
PhysioNet-P300, where the scratch model is itself strong; there, 25\,\% of the
labels bring the pre-trained encoder within 0.9 balanced-accuracy points of
full-data scratch.

\textbf{Missing electrodes.}
Dropping 25\,\% and 50\,\% of the electrodes at test time, without imputation
or retraining, lowers the average score by 2.1 and 4.2 points, so performance
degrades gradually and in proportion to the electrodes removed. With a quarter
of the electrodes missing, KaggleERN balanced accuracy remains 60.4\,\%, still
above EEGPT with the full montage (58.4\,\%), and with half missing it nearly
matches that baseline at 58.3\,\% and still exceeds the scratch model with every
electrode (57.0\,\%). Because absent slots are masked and the remaining channels
keep their name embeddings, the encoder continues to interpret the electrodes it
receives without any adaptation to the reduced montage.

\section{Analysis of Learned Representations}
\label{sec:analysis}
We examine the trained encoders through their spectral responses, scale allocation,
channel embeddings, and reconstructions. Together, these analyses connect the
architecture to the signal structure captured during learning. All analyses use
trained checkpoints without additional optimization.

\subsection{Spatial Structure in Channel Embeddings}
\label{sec:analysis:topology}

\begin{figure}[tbp]
  \centering
  \includegraphics[width=\columnwidth]{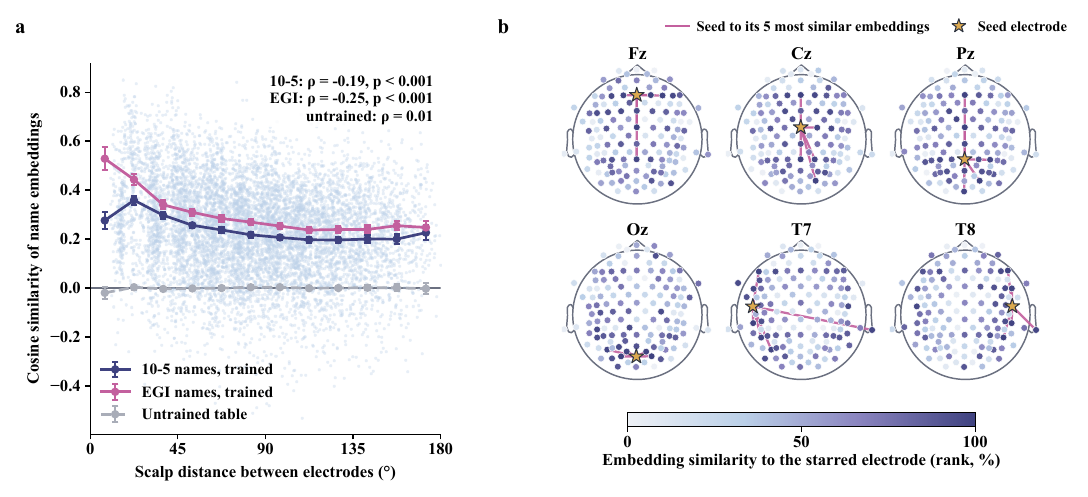}
  \caption{Spatial structure in channel embeddings. (a)~Embedding similarity versus geodesic scalp distance, with an untrained table for comparison. (b)~Electrode-centered similarity maps; links mark the five nearest neighbors.}
  \label{fig:topology}
\end{figure}

Channel identities enter the model only as learned name embeddings
(\eqref{eq:channel}), with no access to electrode coordinates. We test whether
these embeddings nevertheless encode scalp geometry by relating their pairwise
cosine similarities to geodesic electrode distances (Fig.~\ref{fig:topology}a).
For the 130 electrodes of the 10-5 system and the 128 EGI HydroCel positions,
the Spearman correlations are $\rho=-0.19$ and $-0.25$, more than twelve
standard deviations from the permutation-null mean ($p<0.001$ over 2{,}000
draws), whereas the untrained table shows no association ($\rho=0.01$). The
structure is even clearer locally. The five nearest neighbors of each electrode
in embedding space lie at a mean angular distance of 56$^\circ$ on the scalp for
10-5 names and 44$^\circ$ for EGI, compared with 84--88$^\circ$ after name
shuffling, and the electrode-centered maps for Fz, Cz, Pz, Oz, T7, and T8
(Fig.~\ref{fig:topology}b) show similarity concentrated around each reference
site. Because volume conduction makes neighboring electrodes record overlapping
mixtures of the same sources, pre-training on raw signals is enough for the
vocabulary to recover scalp topology from names alone. This learned geometry
also explains the ablation result in Section~\ref{sec:ablation}, where shuffling
channel names costs more than removing channel encoding entirely.

\subsection{Modality-Specific Spectral Adaptation}
\label{sec:analysis:wavelet}

\begin{figure}[tbp]
  \centering
  \includegraphics[width=\columnwidth]{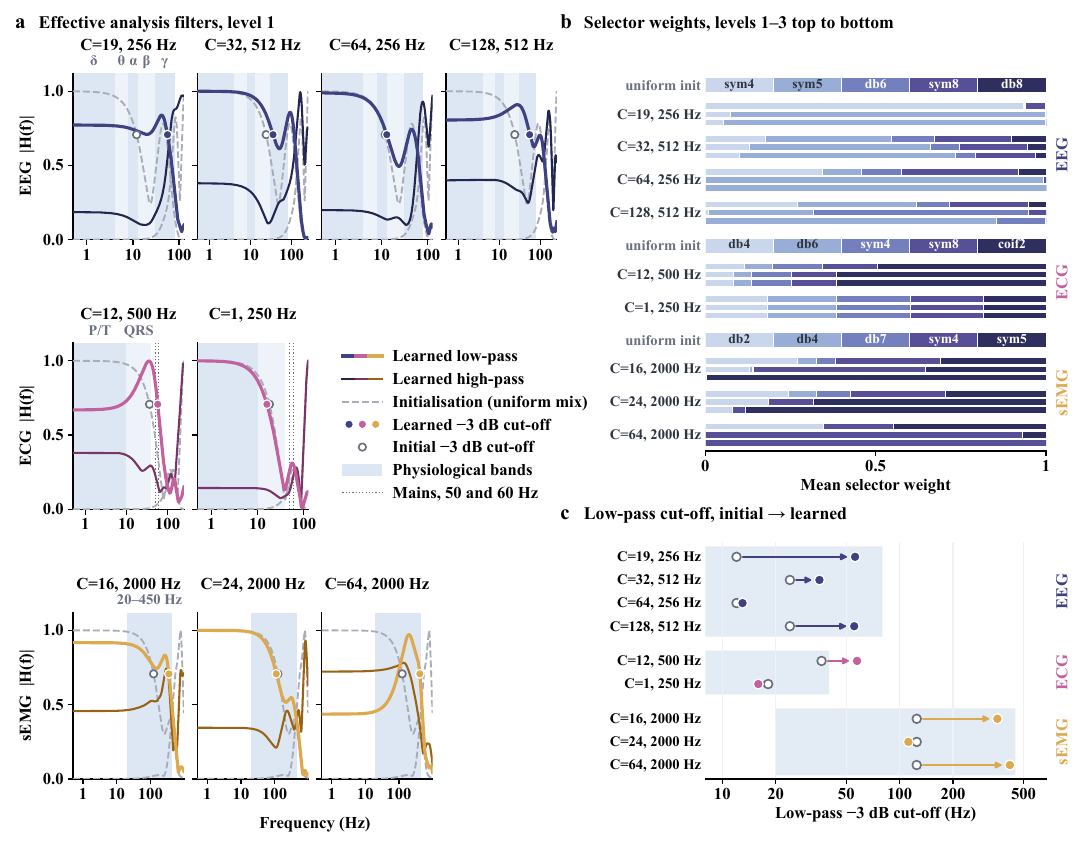}
  \caption{Spectral adaptation of the wavelet front ends. (a)~Level-one analysis responses before and after pre-training, with physiological bands and $-3$\,dB cutoffs. (b)~Selector weights at levels~1--3. (c)~Low-pass cutoff changes.}
  \label{fig:wavelet}
\end{figure}

All front ends start from the same uniformly mixed filter bank, so any
difference after pre-training is learned from the signals themselves
(Fig.~\ref{fig:wavelet}a). The level-one responses move toward the frequency
range that carries each modality's characteristic activity: the QRS range for
12-lead ECG, roughly 100 to 200\,Hz for high-density sEMG, where surface EMG
concentrates its power, and the $\gamma$ range in the high-pass branch of the
19-channel EEG route. The sEMG front ends also widen their low-pass branch, with
$-3$\,dB cutoffs rising from 125\,Hz at initialization to 356--418\,Hz. At
2\,kHz, the initial cutoff would have assigned most of the sEMG band to the
first detail subband; the widened passband keeps it in the approximation that
feeds deeper levels, so the decomposition can resolve muscle activity at
several scales. This adaptation also explains the weak probing result on the
200\,Hz EPN-612 wristband (Section~\ref{sec:results:emg}), whose content lies
largely below the range the pre-trained sEMG filters emphasize.

Basis selection becomes more decisive with depth (Fig.~\ref{fig:wavelet}b). At
level one, every route keeps a mixture over its five candidate wavelets, whereas
at level three the weights concentrate on a single basis within each modality's
bank. Fine resolutions contain heterogeneous transients that benefit from a
blend of wavelet shapes, while the smoother coarse approximations are described
well by one basis, and the selector learns this division without supervision.

\subsection{Scale Allocation Across Sleep Stages}
\label{sec:analysis:fold}

\begin{figure}[tbp]
  \centering
  \includegraphics[width=\columnwidth]{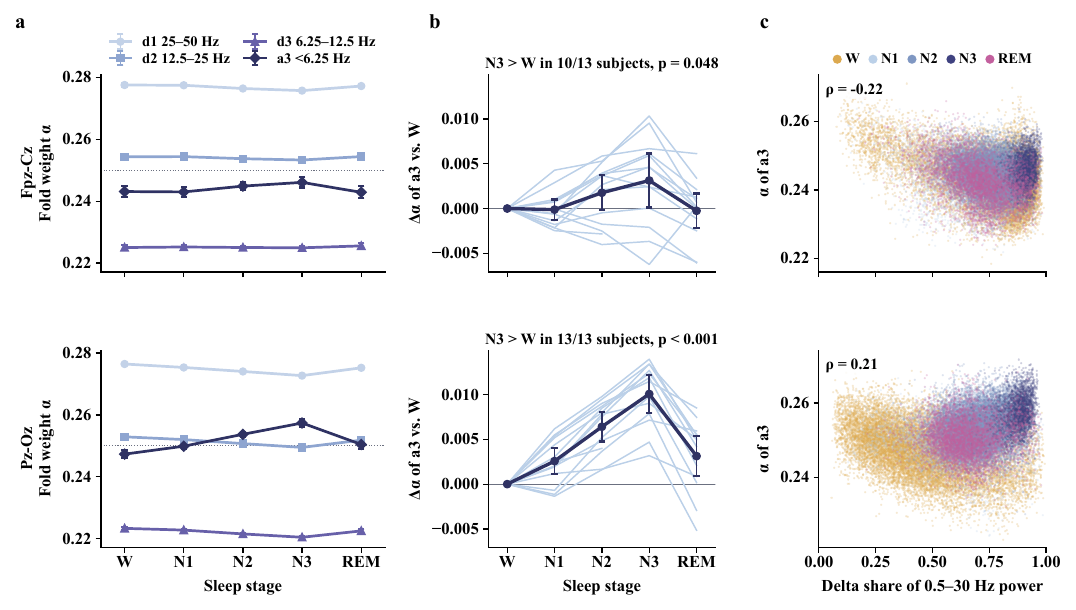}
  \caption{Scale allocation across sleep stages. (a)~Stage-wise folding weights with 95\,\% subject-level confidence intervals. (b)~Within-subject changes relative to wake. (c)~Slowest-scale weight versus $\delta$-band power share. Rows show Fpz-Cz and Pz-Oz.}
  \label{fig:fold_by_stage}
\end{figure}

Sleep staging offers a direct test of whether the folding weights track
physiology, since sleep stages are defined largely by their dominant rhythms.
We analyze the encoder fine-tuned on Sleep-EDFx over 37{,}902 thirty-second
windows from 14 test subjects, grouping the weights by expert-annotated stage
(Fig.~\ref{fig:fold_by_stage}a). The folding network receives only
signal-derived subband statistics and is never told the stage, yet the weight
of the slowest scale, $a_3$ (below 6.25\,Hz), rises steadily from wake to N3 and
returns toward its wake value in REM. This trajectory mirrors the
electrophysiology of sleep: slow-wave activity dominates N3, whereas REM EEG
resembles the low-amplitude, mixed-frequency activity of wake. Within subjects
(Fig.~\ref{fig:fold_by_stage}b), the wake-to-N3 increase holds in all 13
subjects with both stages on Pz-Oz (Wilcoxon $p=2.4\times10^{-4}$) and in 10 of
13 on Fpz-Cz ($p=0.048$), so the effect is consistent across individuals.

The two derivations nonetheless receive different scale mixtures from the same
shared network. At the window level (Fig.~\ref{fig:fold_by_stage}c), the
slowest-scale weight correlates positively with $\delta$-band power share on
Pz-Oz ($\rho=0.21$) but negatively on Fpz-Cz ($\rho=-0.22$). Fpz-Cz also records
ocular activity, whose low-frequency power rises with eye movements during wake
and REM, so $\delta$-band power on this derivation is not a clean marker of
slow-wave sleep. The fold follows the sleep stage on both derivations while
responding differently to their raw low-frequency power, showing that it
allocates scales according to what the low-frequency content represents on
each derivation.

\section{Discussion}
\label{sec:discussion}

\method rests on one organizing principle: everything that depends on how a
signal was acquired is handled before tokenization, and everything that
depends on what it contains is learned afterward by a shared backbone. This is
why a single 30\,M-parameter EEG encoder can learn from 19- to 128-channel
montages and then transfer to a two-channel wearable, and why decomposition
depth becomes a choice about the content of each token instead of the length
of the sequence. The analyses show that the efficiency costs nothing in
physiological meaning: identically initialized front ends settle on the QRS,
100--200\,Hz, and $\gamma$ ranges for their modalities, the folding weights
track sleep stages without stage labels, and the channel vocabulary recovers
scalp topology from electrode names alone, which makes coordinates an
unnecessary input and lets the same identity mechanism cover ECG leads and
sEMG sites.

The experiments also clarify when pre-training matters and how to use it. Its
benefit is smallest where downstream data already cover the test conditions
and largest under distribution shift and label scarcity, as in inter-session
sEMG and KaggleERN, where a tenth of the labels match full-data training from
scratch. Some tasks are served by the frozen representation, whereas subtle
responses such as error-related negativity require adapting the backbone, so
probing first and fine-tuning when needed is the natural recipe. For
multimodal monitoring, RALF shows that per-sample reliability pays off when
artifact-prone ECG must be weighed against EEG and at strict false-alarm
budgets, and its scores fall in proportion to how much signal a corruption
leaves. The corruptions studied here are synthetic; prospective recordings
with natural sensor failures, joint multimodal pre-training, and new routes for
modalities such as photoplethysmography are the next steps.

\section{Conclusion}
\label{sec:conclusion}

\method shows that a biosignal foundation model can accommodate heterogeneous
acquisition without paying for it in sequence length. Route-specific wavelet
front ends and name-based channel embeddings let one backbone per modality
learn from any sensor configuration, dynamic scale folding keeps multiscale
analysis within $C\,S$ tokens, and the resulting EEG, ECG, and sEMG encoders
lead the compared models across sleep staging, event-related potentials,
cardiac diagnosis, and gesture recognition while cutting Transformer memory by
up to an order of magnitude. Reliability-aware fusion extends them to
multimodal seizure detection under missing and corrupted inputs, and new
acquisition setups attach to the released backbones with a new front end
instead of a new model.

\section*{Acknowledgments}
This research was supported by NTU Nanyang Assistant Professor Startup Grant 025661-00021:Efficient Signal Processing Algorithms and Hardware for Accelerated Machine Learning. PhysioWave was partly supported by the EU Horizon Europe projects IntelliMan
(grant agreement 101070136) and HAL4SDV (grant agreement 101139789), the PNRR MUR
project ECS00000033 ECOSISTER.
The authors thank Cineca (Italy)
and CSCS (Switzerland) for GPU computing resources and operational support.

\FloatBarrier
\bibliographystyle{IEEEtran}
\bibliography{refs}

\end{document}